\documentclass[letterpaper]{article} 
\usepackage{aaai25}  
\usepackage{times}  
\usepackage{helvet}  
\usepackage{courier}  
\usepackage[hyphens]{url}  
\usepackage{graphicx} 
\usepackage{natbib}  
\usepackage{caption} 
\usepackage{algorithm}
\usepackage{algorithmic}

\usepackage{amssymb}
\usepackage{amsmath}
\usepackage{multirow}
\usepackage{booktabs}
\usepackage{graphicx}
\usepackage{soul}
\usepackage{array} 
\usepackage[table]{xcolor}
\usepackage{colortbl}
\usepackage{pgfplots}
\usepackage{hyperref}

\usepackage{bm}

\newcommand\scalemath[2]{\scalebox{#1}{\mbox{\ensuremath{\displaystyle #2}}}}

\usepackage{fontawesome}
\usepackage{pgfplots}
\usepackage{pifont}
\usetikzlibrary{positioning,matrix,calc,arrows.meta,decorations.pathreplacing,shapes.geometric}
\definecolor{myred}{RGB}{220,237,217}
\definecolor{myred}{RGB}{246,218,232}
\definecolor{myblue}{RGB}{220,242,248}

\definecolor{mygreen_confidence}{RGB}{77,189,219}
\definecolor{myred_confidence}{RGB}{231,157,194}
\definecolor{myblack}{RGB}{0,0,0}

\definecolor{error_purple}{RGB}{169,158,206}
\definecolor{error_blue}{RGB}{77,189,219}
\definecolor{error_red}{RGB}{231,157,194}

\usepackage{pgfplotstable}

\usepackage{newfloat}
\usepackage{listings}
\DeclareCaptionStyle{ruled}{labelfont=normalfont,labelsep=colon,strut=off} 
\floatstyle{ruled}
\newfloat{listing}{tb}{lst}{}
\floatname{listing}{Listing}
\title{RoVRM: A Robust Visual Reward Model Optimized via Auxiliary \\ Textual Preference Data}

\author{
    Chenglong Wang\textsuperscript{\rm 1}, 
    Yang Gan\textsuperscript{\rm 1}, 
    Yifu Huo\textsuperscript{\rm 1}, 
    Yongyu Mu\textsuperscript{\rm 1}, 
    Murun Yang\textsuperscript{\rm 1}, 
    Qiaozhi He\textsuperscript{\rm 1}, \\ 
    Tong Xiao\textsuperscript{\rm 1,2}\footnote{Corresponding author.}, 
    Chunliang Zhang\textsuperscript{\rm 1,2}, 
    Tongran Liu\textsuperscript{\rm 3},
    Quan Du\textsuperscript{\rm 2},
    Di Yang\textsuperscript{\rm 2}
    and 
    Jingbo Zhu\textsuperscript{\rm 1,2}
}
\affiliations{
    \textsuperscript{\rm 1} School of Computer Science and Engineering, Northeastern University, Shenyang, China \\
    \textsuperscript{\rm 2} NiuTrans Research, Shenyang, China \\
    \textsuperscript{\rm 3} CAS Key Laboratory of Behavioral Science, Institute of Psychology, CAS, Beijing, China \\
    {\{clwang1119, zzhu8250\}@gmail.com},
    {\{xiaotong, zhujingbo\}@mail.neu.edu.cn}
}

\usepackage{bibentry}

\begin{document}

\maketitle

\begin{abstract}
Large vision-language models (LVLMs) often fail to align with human preferences, leading to issues like generating misleading content without proper visual context (also known as \textit{hallucination}).
A promising solution to this problem is using human-preference alignment techniques, such as best-of-$n$ sampling and reinforcement learning.
However, these techniques face the difficulty arising from the scarcity of visual preference data, which is required to train a visual reward model (VRM).
In this work, we continue the line of research.
We present a \textbf{Ro}bust \textbf{V}isual \textbf{R}eward \textbf{M}odel (RoVRM) which improves human-preference alignment for LVLMs.
RoVRM leverages auxiliary textual preference data through a three-phase progressive training and optimal transport-based preference data selection to effectively mitigate the scarcity of visual preference data.
We experiment with RoVRM on the commonly used vision-language tasks based on the LLaVA-1.5-7B and -13B models.
Experimental results demonstrate that RoVRM consistently outperforms traditional VRMs. 
Furthermore, our three-phase progressive training and preference data selection approaches can yield consistent performance gains over ranking-based alignment techniques, such as direct preference optimization.

\end{abstract}

\section{Introduction}
Large language models (LLMs) have demonstrated remarkable capabilities across various natural language processing tasks \cite{stiennon2020learning,ouyang2022training}.
Recent works tend to fine-tune LLMs using specialized visual instruction tuning datasets, leading to the emergence of powerful large vision-language models (LVLMs) \cite{liu2024improved,lin2024vila,huang2024language}.
Despite these advancements, current LVLMs are not well-aligned with human preferences. 
A glaring problem is that LVLMs sometimes generate misleading content without anchoring to the given visual context (also known as \textit{hallucination}) \cite{leng2024mitigating}.
For instance, as illustrated in Figure \ref{fig:main-image}, an LVLM incorrectly identifies a “pitaya” in an image of mangosteens due to their visual similarity.

Two predominant research approaches aim to address this problem. 
The first approach focuses on generating richer and higher-quality visual instruction data \cite{li2023stablellava,liu2023aligning,liu2024mm}, \textit{i.e.}, annotating rich instruction samples on images of mangosteens to enable LVLMs to identify them more accurately.
In contrast, a more sophisticated approach is applying human-preference alignment techniques, including best-of-$n$ sampling and reinforcement learning (RL), which can efficiently align models with human preferences on various tasks by optimizing against a reward model without instruction samples.
However, applying these alignment techniques to LVLMs is not a low-hanging fruit.
It typically faces the difficulty of training a visual reward model (VRM) due to the scarcity of high-quality visual preference data \cite{sun2023aligning,yu2024rlhf,zhou2024aligning}.

This work is motivated by a simple idea: human preferences are well-captured by text and these preferences can be transferred across different modalities. 
In this way, we can make use of rich, high-quality textual preference data in training VRMs. 
Building on this idea, we present a  \underline{\textbf{Ro}}bust \underline{\textbf{V}}isual \underline{\textbf{R}}eward \underline{\textbf{M}}odel (RoVRM), which can improve human-preference alignment for LVLMs in two ways.
For one, we propose a three-phase progressive training approach to gradually bridge the task and modality gaps between textual and visual preference data, which can take full advantage of auxiliary textual preference data to improve the robustness of RoVRM.
Furthermore, considering the conflict in preferences \cite{coste2023reward, eisenstein2023helping}, leveraging textual preference data poses a problem: not all data is beneficial for training the RoVRM.
Addressing this problem, we propose an optimal transport-based preference data selection approach.
This approach can select textual preference data that better aligns with the vision-language task preferences, thereby improving the efficacy of the RoVRM training process.
To the best of our knowledge, we are the first to investigate the integration of preferences from different modalities.

Through experiments on five vision-language tasks, we aim to comprehensively evaluate RoVRM using two commonly used human-preference alignment techniques: best-of-$n$ sampling and RL. 
Our results demonstrate improved performance in each task when aligned with reward signals from RoVRM, as confirmed by both automatic and human evaluations.
Notably, when performing best-of-$n$ sampling on the LLaVA-1.5-7B model, RoVRM outperforms a traditional VRM by 8.4 points on the LLaVA-Bench benchmark.

As another bonus, our three-phase progressive training and preference data selection can be seamlessly integrated with arbitrary ranking-based alignment techniques, such as direct preference optimization (DPO) \cite{rafailov2024direct}, SimPO \cite{meng2024simpo}, and ORPO \cite{hong2024orpo}.
For instance, on the LLaVA-1.5-13B model, integrating with DPO results in an additional improvement of 17.82 points on the MM-Instruct benchmark compared to standard DPO.



\section{Related Work}
In recent years, LVLMs have served as the primary backbone for vision-language tasks \cite{achiam2023gpt,ai2023fuyu}.
Aligning LVLMs with human preferences is effective in gaining more performance \cite{liu2023aligning,wang2024mdpo}.
However, in this process, they only used visual preference data and never leveraged the textual preference data that exists in abundance.

\subsubsection{Large Vision-Language Models}
Inspired by the success of LLMs such as GPTs \cite{brown2020language,ouyang2022training} and LLaMA \cite{touvron2023llama}, researchers have been aiming to develop LVLMs. 
The basic idea is to augment LLMs with visual inputs (\textit{e.g.,} images) to provide an interface for vision-language tasks \cite{alayrac2022flamingo,awadalla2023openflamingo,aiello2023jointly}.
Recent works on LVLMs could be classified into two groups.
The first group focused on integrating visual information into LLMs \cite{chen2023minigpt,liu2024improved,wang2024visionllm}.
For example, \citet{liu2024visual} constructed a large amount of visual instruction data to pre-train the visual projection layer. 
\citet{lin2024vila} further investigated the effective pre-training design options to augment LVLMs.
The second group that has attracted attention commonly aimed to improve the consistency of output text and visual content, particularly addressing the problem of hallucination \cite{zhou2023analyzing,leng2024mitigating,gunjal2024detecting,huang2024opera,favero2024multi}.
This work belongs to the latter, where our RoVRM can improve the consistency of output text and visual content.

\subsubsection{Human-Preference Alignment for LVLMs}
Reinforcement learning with human feedback (RLHF) has been shown to effectively align LLM behaviors with human preferences \cite{stiennon2020learning,ouyang2022training}. 
Several works have improved RLHF by using fine-grained reward models \cite{wu2024fine}, reward model ensembles \cite{coste2023reward}, and direct preference optimization objectives \cite{rafailov2024direct}. 
Additionally, some works focused on generating large, high-quality textual preference datasets to further improve RLHF in LLMs \cite{cui2023ultrafeedback,dubois2024alpacafarm}.
In the context of LVLMs, existing works mainly focused on the adaptation of the human-preference alignment techniques \cite{sun2023aligning,li2023silkie,yu2024rlhf}.
A significant challenge here was the scarcity of visual preference data.
To address this challenge, many efforts have been made to create visual preference data, including collecting human preferences \cite{sun2023aligning}, and acquiring preferences from a strong LVLM \cite{li2023silkie,yu2024rlaif}.
Different from these works, we investigate how to leverage rich, high-quality textual preference data to offset the scarcity of visual preference data.

\section{Our Method}
We first review the preliminaries of the human-preference alignment training for language models.
Then, we present the three-phase progressive training for use with RoVRM. 
Last, we introduce the proposed preference data selection.

\subsection{Preliminaries}
\subsubsection{Reinforcement Learning with Human Feedback}
RLHF is a key technique for aligning language models with human preferences.
It typically consists of two main steps: 1) training a reward model (also known as preference model) from preference data, and 2) using an RL algorithm, such as PPO \cite{schulman2017proximal}, to maximize the reward.
In step 1, we usually employ the Bradley-Terry model \cite{bradley1952rank}.
When the preference data existed in a comparison pair, the loss function can be written as:
\begin{eqnarray}
    \mathcal{L}_{reward}=-\log (\sigma (r_{\theta}(x,y_{w})-r_{\theta}(x,y_{l})))
    \label{eq:reward_training_1}
\end{eqnarray}
where $\sigma$ is the Sigmoid activation function, $r(\cdot)$ is a reward model and $\theta$ is its parameters.
$y_{w}$ and $y_{l}$ are two different responses for the human prompt $x$, where $y_{w}$ is more preferred than $y_{l}$.
When dealing with multiple responses more than two, we can induce $\mathcal{L}_{reward}$ based on the more general Plackett-Luce model \cite{luce2005individual}:
\begin{eqnarray}
    \scalemath{0.95}{
    \mathcal{L}_{reward}= -\sum^{k}_{i=1} \log \frac{\exp\left({r_{\theta}(x,y_{i})}\right)}{\sum_{j=i}^{k} \exp\left({r_{\theta}(x,y_{j})}\right)}}
    \label{eq:reward_training_2}
\end{eqnarray}
where $k$ denotes the number of responses.
These responses are ranked by the defined preferences: 
$\left(y_{1} \succ \cdots \succ y_{k} | x\right)$, where $y_{1}$ is the best while $y_{k}$ is the worst.
In step 2, the reward signals produced by the trained reward model are instrumental in adjusting the parameters of the language models.
Thus, the alignment of the language model is significantly influenced by how well the reward model is trained.

\begin{figure*}
    \centering
    \includegraphics[width=0.95\linewidth]{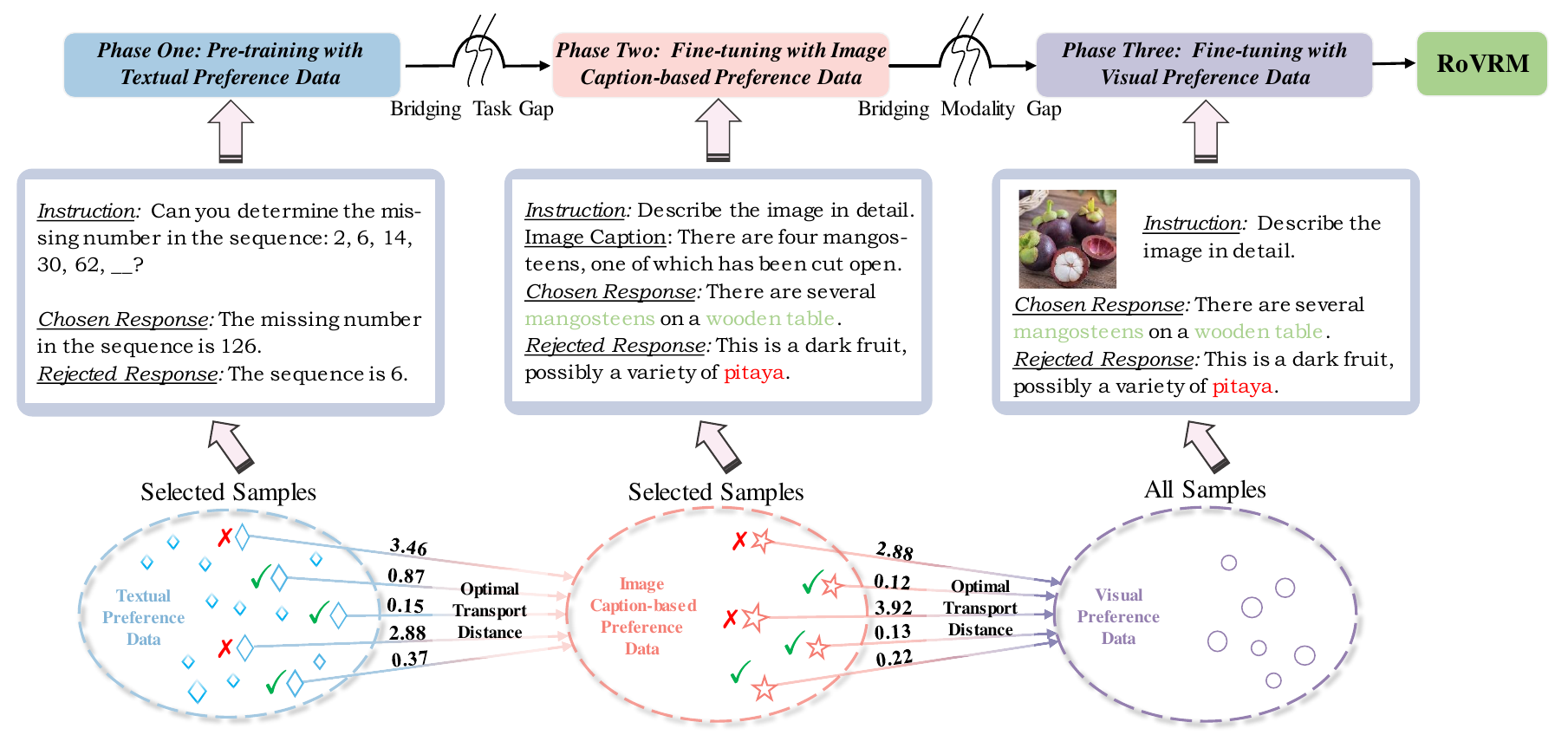}
    \vspace{-2mm}
    \caption{
    We propose three-phase progressive training and optimal transport-based preference data selection approaches to train RoVRM.
    For three-phase progressive training, we take full advantage of textual preference data to compensate for the limited availability of visual preference data. 
    Using this preference selection, samples for phases one and two are selected based on those selected for the subsequent phase. 
    \textcolor{green}{\checkmark} denotes a selected sample, while \textcolor{red}{\ding{55}} denotes one that is not selected.
    }
    \vspace{-5mm}
    \label{fig:main-image}
\end{figure*}

\subsubsection{Direct Preference Optimization}
To bypass the complex RL procedure, \citet{rafailov2024direct} proposed the direct preference optimization (DPO) which employs a reward model training objective to maximize rewards.
It gives a new loss function:
\begin{eqnarray}
\scalemath{0.90}{
\mathcal{L}_{\mathrm{DPO}}=-\log \sigma [ \beta \log(\frac{p_{\theta'}(y_{w}|x)}{p_{\theta'_{old}}(y_{w}|x)})-\beta \log(\frac{p_{\theta'}(y_{l}|x)}{p_{\theta'_{old}}(y_{l}|x)})]}
\end{eqnarray}
where $\theta'$ denotes the parameters of the language model, $\theta'_{old}$ denotes the parameters of the language model trained via supervised fine-tuning, $\beta$ denotes a scaling factor, and $\sigma$ denotes a Sigmoid function.

\subsubsection{Best-of-$n$ Sampling}
Best-of-$n$ sampling (also known as re-ranking) refers to reordering or reevaluating a set of candidate responses sampled from a trained model \cite{lee2021discriminative,fernandes2022quality}.
Given a set $\mathcal{Y}$ of $n$ candidate responses for $x$, we can also use the best-of-$n$ sampling approach to maximize the reward, thereby aligning the response with human preferences.
Typically, we employ the reward model to score the candidate responses and select a final response that has a maximum reward score.

We can notice that when applying these alignment training methods to LVLMs, sufficient visual preference data is required either to train a VRM or to perform DPO training. 
However, in practice, visual preference data is often insufficient and expensive to acquire.

\subsection{A Robust Visual Reward Model}
We aim to provide a RoVRM for human-preference alignment in LVLMs.
The overview of training RoVRM is depicted in Figure \ref{fig:main-image}.
As shown in the figure, we present a three-phase progressive training and preference data selection to improve the robustness of RoVRM.

\subsubsection{Three-Phase Progressive Training}
In response to the scarcity of visual preference data, we propose a three-phase progressive training approach that effectively solves this issue.
Phase one is to conduct preference pre-training using a large amount of textual preference data. 
This phase can help our RoVRM to pre-learn general preferences.
Ideally, the RoVRM would inherit these general preferences when processing vision-language tasks.
However, this faces two serious obstacles: \textit{task gap} and \textit{modality gap}, which prevent these preferences from being directly applicable to vision-language tasks (see experiments in Figure \ref{fig:few_shot}).
Here, we design phases two and three to bridge these gaps progressively. 
Phase two is to bridge the task gap by constructing vision-language preference data based on image captions and fine-tuning the RoVRM.
Specifically, we use image captions to replace the images for visual preference data, \textit{i.e.,} changing the human prompt $x$=[\texttt{Instruction}; \texttt{Image}] to $x$=[\texttt{Instruction}; \texttt{Image Caption}] in Eqs. \ref{eq:reward_training_1} and \ref{eq:reward_training_2}.
Building on phase two, phase three is to bridge the modality gap by using the visual preference data to continue fine-tuning the RoVRM with a visual projector.
Compared to training a VRM directly with visual preference data, this three-phase training process incurs additional time costs due to an extra preference training session. 
However, it can leverage auxiliary textual preference data to improve robustness and respond to the scarcity of visual preference data.
Furthermore, although pre-training followed by fine-tuning is widely used in machine learning \cite{devlin2018bert,liu2019roberta,xiao2023introduction}, our approach is the first to demonstrate the feasibility of optimizing a VRM through this paradigm.

\subsubsection{Preference Data Selection}
Not all preference data aligns with the preferences used in subsequent phases, and conflicts may arise. 
Thus, during each training phase, we expect to employ samples that more closely align with the preferences contained in the data for the next phase.
To achieve this, we propose an optimal transport-based preference data selection approach. 
We apply this approach to perform preference data selection for phases one and two, based on the preference data used in the next phase.
For instance, in phase one, following \citet{xia2024less}'s work, we first extract gradient features for all samples in the textual preference dataset $\mathcal{D}_\mathrm{T}=\{s_1^{t},s_2^{t},\cdots,s_m^{t}\}$. Based on these features, we compute the distance score between each sample in $\mathcal{D}_\mathrm{T}$ and the image caption-based preference dataset $\mathcal{D}_\mathrm{C}=\{s_1^{c},s_2^{c},\cdots,s_n^{c}\}$ using optimal transport.
The details are described as follows.

\textit{Gradient Feature.}
\citet{xia2024less} construct gradient features for each sample of general supervised fine-tuning data to select the data that more effectively improves the specific downstream task.
Here, using these gradient features, we conduct the preference data selection.
Specifically, we firstly use LoRA \cite{hu2021lora} to efficiently perform a warmup reward model training with a small subset of preference data $\mathcal{D}_{\mathrm{Warmup}}$, where $\mathcal{D}_{\mathrm{Warmup}}$ is a subset extracted randomly from $\mathcal{D}_{\mathrm{T}}\cup\mathcal{D}_{\mathrm{C}}$.
Then, we extract the gradient features for each preference sample in $\mathcal{D}_{\mathrm{T}}$ and $\mathcal{D}_{\mathrm{C}}$ through the forward and backpropagating on the warmed-up reward model:
\begin{eqnarray}
    g=\mathrm{RP}(\nabla\mathcal{L}_{reward}(s;\theta_{warmup}))
\end{eqnarray}
where $g$ is the gradient feature of the preference sample $s$ and $\theta_{warmup}$ is the parameters of the warmed-up reward model.
$\mathrm{RP}(\cdot)$ is a random projection \cite{xie2017survey} that reduces the dimensionality of gradient features.

\textit{Optimal Transport-based Distance.}
Unlike the \citet{xia2024less} who use the cosine similarity to compute sample distance scores, we use optimal transport \cite{villani2009optimal}, endowed with the capability to compute the distance transferring an arbitrary data feature to a specific data feature \cite{gurumoorthy2021spot,kang2024get}.
Our motivation is to gather preference data for easy integration into the next training phase. To reduce computational overhead, we select a representative subset $\mathcal{D}_\mathrm{SubC}$ from $\mathcal{D}_\mathrm{C}$. This subset approximates the distance computation for the entire dataset $\mathcal{D}_\mathrm{C}$ when selecting samples from $\mathcal{D}_\mathrm{T}$.
We define the distance score of $i$-th sample in $\mathcal{D}_{\mathrm{T}}$ by:
\begin{eqnarray}
    c_{i} = \frac{1}{|\mathcal{D}_\mathrm{SubC}|}\sum_{j=1}^{|\mathcal{D}_\mathrm{SubC}|}\mathrm{OT}(g_{i}^{t}, g_{j}^{c})
    \label{eq:distance_score}
\end{eqnarray}
where $g_{i}^{t}$ and $g_{j}^{c}$ denote the gradient features for the preference samples $s_{i}^{t}$ and $s_{j}^{c}$, respectively.
$\mathrm{OT}(\cdot)$ denotes the function of computing the transfer distance.
Given gradient features $g^{t}_{i}$, $g^{c}_{j}$ over a gradient space $\mathcal{Z}$, the optimal transport-based transfer distance can be defined as:
\begin{eqnarray}
    \mathrm{OT}(g^{t}_{i}, g^{c}_{j}) := \min_{\gamma \in \Gamma(g^{t}_{i}, g^{c}_{j})} \int_{\mathcal{Z}^2} C(z, z') \, d\gamma(z, z')  
\end{eqnarray}
where $C(\cdot)$ denotes a symmetric positive-definite cost function, and $\Gamma(g^{t}_{i}, g^{c}_{j})$ denotes a collection of couplings between two gradients $g^{t}_{i}$ and $g^{c}_{j}$.
Here, we utilize $L_2$-norm as the cost function and define the sum of the solved $\gamma$ as the distance score.
A lower distance score indicates that the textual preference sample has preferences more easily transferable to the vision-language task. 
Our implementation of optimal transport solvers is done using Python Optimal Transport (\texttt{POT})\footnote{\url{https://pythonot.github.io/index.html}}.
While optimal transport distance has been used in data selection before \cite{kang2024get}, this is the first application to preference data selection.

To ensure that the ultimate goal of selecting preference data is to transfer preferences from textual preference data to vision-language tasks, we start by selecting image caption-based preference data for phase two. 
Next, we choose the textual preference data for phase one based on the preference data selected in phase two.

\section{Experiments}
We evaluated our RoVRM on the commonly used vision-language tasks based on the best-of-$n$ sampling and reinforcement learning (RL).
We also evaluated our approaches to direct preference optimization (DPO).

\subsection{Experimental Setups}
\subsubsection{Datasets}
The datasets used in this work are as follows:
\begin{itemize}
    \item \textit{Textual Preference Dataset}:
    We used UltraFeedback \cite{cui2023ultrafeedback}, a large-scale, high-quality, and diversified preference dataset, as our textual preference dataset.
    It comprises 64k instructions, each with 4 responses, leading to over 340k comparison preference pairs.
    
    \item \textit{Image Caption-based Preference Dataset}:
    We constructed an image caption-based preference dataset to bridge the task gap. 
    Specifically, we employed \texttt{GPT-4o-mini} to generate detailed image captions that replace the visual content in our preference data. 
    Note that when the image is present in the COCO caption dataset\footnote{\url{https://huggingface.co/datasets/lmms-lab/COCO-Caption2017}}, we used the human-annotated captions directly.
    
    \item \textit{Visual Preference Dataset}:
    We employed the visual preference dataset from RLAIF-V \cite{yu2024rlaif}, which consists of about 83k comparison preference pairs.
    To our knowledge, it is the largest scale open source preference dataset in computer vision.
    
    \item \textit{RL Training}:
    We sampled 50k instructions from LLaVA-Instruct-150K \cite{liu2024visual} for training. 
\end{itemize}

\vspace{-2mm}

\begin{table*}[t]
    \centering
    \scalebox{0.85}{

\begin{tabular}{llllllllll}
\toprule[1.0pt]
\multicolumn{1}{l}{\multirow{2}{*}{Method}} & \multicolumn{1}{c}{\multirow{2}{*}{\#Param}} & \multicolumn{2}{c}{MMHalBench} & \multicolumn{2}{c}{Object HalBench} & \multicolumn{2}{c}{AMBER} & LLaVA$\mathrm{^W}$ & \multicolumn{1}{c}{MMIns} \\ \cmidrule(l){3-4} \cmidrule(l){5-6} \cmidrule(l){7-8} \cmidrule(l){9-9} \cmidrule(l){10-10} 

\multicolumn{1}{c}{}  & \multicolumn{1}{c}{} & Score $\uparrow$ & HalRate $\downarrow$ & Resp. $\downarrow$ & Ment. $\downarrow$ & Cover. $\uparrow$ & HalRate $\downarrow$ & Score $\uparrow$ & WinRate $\uparrow$ \\ \midrule
   GPT-4V \cite{achiam2023gpt}$^\sharp$ &  -& 3.49 & 28.1 &13.6   &7.3  &67.1  &30.7   & 93.1   & 100.00 \\
   GPT-4o      &  -&3.58  &26.0 &10.7 &5.2 & 64.0 &24.9   &126.4  & 100.00\\
   GPT-4o-mini &  -&3.02  &29.2 &9.1 &5.4 & 58.4 & 18.5 &130.9 & 100.00 \\ \midrule
   Qwen-VL-Chat \cite{bai2023qwen}$^\sharp$ & 10B & 2.76 &38.5 & 40.4 & 20.7 &53.2 &31.0  & 71.9 & 73.58\\
   OmniLMM \cite{hu2023large}$^\dagger$    &  12B&  3.14&  36.5 & 12.2 & 6.2 & - & -& 72.7 & -\\
   MiniGemini \cite{li2024mini}$^\dagger$  &  34B&  3.08&  38.5 & 14.5 & 8.0 & - & - &79.2  & -\\
   LLaVA-NeXT \cite{liu2024llavanext}$^\dagger$ &  34B&  3.31&  34.4 &12.6 &6.4 &63.2 &43.6 & 77.7 &93.83 \\ 
   LLaVA-1.5-7B \cite{liu2024improved}$^\sharp$$^\ddagger$ & 7B & 2.36 & 51.0 & 53.6 & 25.2  & 51.8 & 34.7 & 65.4 & - \\
   LLaVA-1.5-13B \cite{liu2024improved}$^\sharp$ & 13B &2.42 &- &46.3 &22.6 &-  &-  & 72.5 & -  \\  \midrule
   LURE \cite{zhou2023analyzing}$^\dagger$ &   7B&  1.64&  60.4 & 27.7 & 17.3 &- &- & 36.9 & - \\  
   HA-DPO \cite{zhao2023beyond}$^\dagger$ &   7B&  1.98& 60.4& 39.9 &19.9 & 49.5 & 29.1 & 60.3 & - \\
   VCD \cite{leng2024mitigating}$^\dagger$&   7B&  2.12&  54.2&  48.8&  24.3 & 51.5 & 39.0 & 65.8 & 42.56   \\
   Silkie \cite{li2023silkie} $^\dagger$ & 10B& 3.19 &32.3 & 27.1 & 13.4 & 56.0 & 28.4 & 73.2 & 63.64 \\
   LLaVA-RLHF \cite{sun2023aligning}$^\dagger$ & 13B & 2.02 & 62.5 & 38.1 & 18.9& 52.0 & 39.2&  61.5 & 74.24 \\
   RLHF-V \cite{yu2024rlhf}$^\dagger$ & 13B & 2.45 & 51.0 & 12.2 & 7.5 & - & -  & 51.4 & -  \\
   \midrule
    \multicolumn{10}{l}{\textbf{\textit{Best-of-$n$ Sampling}}}   \\ \midrule
   LLaVA-1.5-7B        &   7B&2.12  &55.0  &50.3  &29.0  &50.3  &37.1  &  66.7 & 46.16 \\
\ \ \ \ \ +VRM-Vanilla &   7B&2.39  &47.9  &35.3 &21.2  &50.8 &29.0 &  73.6 & 57.69   \\ \rowcolor{gray!20}
\ \ \ \ \ +RoVRM-Random &  7B&2.52  &43.8  &32.7  &18.9    &51.7   &26.9   &77.2  & 58.49  \\ \rowcolor{gray!20}
\ \ \ \ \ +RoVRM &  7B&\bf{2.68}  &\bf{40.6} &\bf{30.4}  &\bf{16.8}  &\bf53.2   &\bf23.9   &\bf82.0  &\bf{61.91}    \\ \midrule
   LLaVA-1.5-13B   & 13B &2.30 &53.8 &49.0  &25.8 &50.6 &37.2 &75.6 & 50.00    \\
\ \ \ \ \ +VRM-Vanilla  & 13B &2.41 &51.0  &32.7  &16.7  &51.4 &26.6   &84.0  &73.08   \\ \rowcolor{gray!20}
\ \ \ \ \ +RoVRM-Random &  13B&2.43  &48.3  &29.0  &15.7  &51.9   &25.7   &86.4  &74.42   \\ \rowcolor{gray!20}
\ \ \ \ \ +RoVRM        &  13B&\bf{2.57}  &\bf{47.3}  &\bf{26.8}  &\bf{13.1}  &\bf{53.6}   &\bf{22.8}   &\bf{89.8}  &\bf{78.75}    \\ \midrule
\multicolumn{10}{l}{\textbf{\textit{Reinforcement Learning}}}  \\ \cmidrule(l){1-10}
   LLaVA-1.5-7B   &   7B&2.12  &55.0  &50.3  &29.0  &\bf{50.3}  &37.1  &  66.7 & 46.16 \\
\ \ \ \ \ +VRM-Vanilla &  7B&2.17  &53.2  &37.7 &26.0 &49.1  &29.1  &72.8  &51.11   \\ \rowcolor{gray!20}
\ \ \ \ \ +RoVRM-Random &  7B&2.21  &50.8  &31.3 &22.0 &48.7  &24.3&74.2  &54.35   \\ \rowcolor{gray!20}
\ \ \ \ \ +RoVRM & 7B  &\bf{2.36}  &\bf{48.9}  &\bf{27.0}  &\bf{16.3}  &48.2   &\bf{23.4}  &\bf{78.3}  &\bf{58.69}    \\\midrule
   LLaVA-1.5-13B   & 13B &2.30 &53.8 &49.0  &25.8 &\bf{50.6} &37.2 &75.6 & 50.00    \\ 
\ \ \ \ \ +VRM-Vanilla  & 13B&2.49  &50.0  &27.8  &16.1  &41.1   &23.2   &78.2  &52.63   \\ \rowcolor{gray!20}
\ \ \ \ \ +RoVRM-Random & 13B&2.34  &47.9  &31.7  &15.3  &48.6 &21.0  &79.5 &60.53   \\ \rowcolor{gray!20}
\ \ \ \ \ +RoVRM        & 13B&\bf{2.57}  &\bf{43.8}  &\bf{25.0}  &\bf{13.2}  &47.7 &\bf{19.5}   &\bf{81.7}  &\bf{65.79}   \\
\bottomrule[1.0pt]
\end{tabular}}
    \vspace{-1mm}
    \caption{
    Experimental results on different vision-language tasks.
    The best results for each group are in \textbf{bold}.
    Results marked with $\dagger$ for MMHalBench, Object HalBench, and LLaVA$^\mathrm{W}$ are from \citet{yu2024rlaif}. 
    Results marked with $\sharp$ for AMBER are from \citet{wang2024mdpo}. 
    Results marked with $\ddagger$ for LLaVA$^\mathrm{W}$ are from \citet{liu2024improved}. 
    The other baseline results are obtained by testing this available model or using the provided API.
    }
    \vspace{-5mm}
    \label{tab:main_experiments}
\end{table*}

\subsubsection{Settings}
For training RoVRM, we used the LLaVA-1.5-7B model to initialize the visual reward model.
The learning rates for the three-phase progressive training were set to 2e-5 for phase one, and 1e-6 for phases two and three. 
For optimal transport-based preference data selection, we used 5k samples to warm up the VRM, consisting of 2k samples from the dataset to be selected and 3k samples from the target preference dataset. 
The representative subset size was set to 5k samples.
For best-of-$n$ sampling and RL training, we employed the LLaVA-1.5-7B as the initial model.
In the process of best-of-$n$ sampling, we set the sampling size to 8. 
We also tested other sampling sizes in Figure \ref{fig:diff_sampling_sizes_BoS}.
More training settings are shown in Appendix A.

\subsubsection{Evaluation}
We evaluated the RoVRM in two key aspects: trustworthiness, which denotes the level of hallucination, and helpfulness, which reflects overall interaction capability. Trustworthiness was evaluated using three benchmarks: Object HalBench \cite{rohrbach2018object}, MMHal-Bench \cite{sun2023aligning}, and AMBER \cite{wang2023llm}.
We reported the response-level (\textbf{Resp.}) and mention-level (\textbf{Ment.}) hallucination rates in the Object HalBench. 
GPT-4 was employed to evaluate the response-level hallucination rate (\textbf{HalRate}) and informativeness score (\textbf{Score}) on the MMHalBench. 
Additionally, we provided the object coverage (\textbf{Cover.}) and hallucination rate metrics for AMBER.
To assess helpfulness, we used two benchmarks: MM-Instruct \cite{liu2024mm} and LLaVA-Bench (In-the-Wild) \cite{liu2024visual}. 
GPT-4, following the settings in \texttt{lmms-eval}\footnote{\url{https://github.com/EvolvingLMMs-Lab/lmms-eval}}, was used to score responses in LLaVA-Bench. 
For MM-Instruct, responses from LLaVA-1.5-13B were used as a baseline, and we computed the win rate (\textbf{WinRate}) as per \citet{liu2024mm}. 
Additionally, a human evaluation was conducted to validate further RoVRM's effectiveness.

\subsubsection{Baselines}
Our baseline were the \textbf{LLaVA-1.5-7B} and \textbf{-13B} models without human-preference alignment.
We also compared with other general LVLMs, such as \textbf{GPTs} \cite{achiam2023gpt} and \textbf{Qwen-VL-Chat} \cite{bai2023qwen}.
Furthermore, we compared RoVRM with commonly used methods to solve the hallucination, including \textbf{LURE} \cite{zhou2023analyzing}, \textbf{HA-DPO} \cite{zhao2023beyond}, and others.
The traditional VRM training was also our baseline, where we optimized a VRM only using our visual preference dataset (\textbf{VRM-Vanilla}).
To evaluate the effectiveness of optimal transport in preference data selection, we chose \textbf{RoVRM-Random} as a baseline.
In RoVRM-Random, we randomly selected samples during the preference data selection.

\begin{table}[t]
    \centering
    \scalebox{0.82}{
    \begin{tabular}{lcccc}
\toprule[1.1pt]
\multirow{2}{*}{Method}  
& \multicolumn{2}{c}{AMBER}                  
& LLaVA$^\mathrm{{W}}$
& MMIns
\\ \cmidrule(r){2-3} \cmidrule(r){4-4} \cmidrule(r){5-5}
&  Cover. $\uparrow$ & HalRate $\downarrow$ & Score $\uparrow$  & WinRate $\uparrow$  \\ \midrule
LLaVA-1.5-7B  &50.3  &37.1 &66.7 &46.16          \\ \midrule
\multicolumn{5}{l}{\textbf{\textit{Best-of-$n$ Sampling}}} \\ \midrule
RoVRM &  \bf{53.2}& \bf{23.9} & \bf{82.0} & \bf{61.91}             \\
\ \ \ \  \ w/o PDS          &52.4 &25.1 & 80.6  & 61.36  \\
\ \ \ \  \ w/o TPT-One      &51.0 &26.7 & 71.3  & 59.52 \\
\ \ \ \  \ w/o TPT-Two      &51.8 &24.9 & 78.0  & 54.76 \\ \midrule
\multicolumn{5}{l}{\textbf{\textit{Reinforcement Learning}}} \\ \midrule
RoVRM     &\bf{48.2}   &\bf{23.4}  &\bf{78.3}  &\bf{58.69}       \\
\ \ \ \  \ w/o PDS       &46.2   &32.2   &75.2  &53.70   \\
\ \ \ \  \ w/o TPT-One   &44.3   &35.0   &73.0  &51.85   \\
\ \ \ \  \ w/o TPT-Two   &47.5   &28.2   &76.1  &55.56    \\
\bottomrule[1.1pt]
\end{tabular}}
    \caption{
    The suffixes “-One” and “-Two” denote the removal of phases one and two, respectively, in the three-phase progressive training approach.
    “w/o PDS” denotes that all data is used for each training phase without employing preference data selection.
    \textbf{PDS}: preference data selection;
    \textbf{TPT}: three-phase progressive training.
    }
    \vspace{-5mm}
    \label{tab:ablation_study}
\end{table}

\subsection{Experimental Results}
\subsubsection{Results of Best-of-$n$ Sampling}
Table \ref{tab:main_experiments} summarizes the performance of our RoVRM on the best-of-$n$ sampling.
On all vision-language tasks, RoVRM consistently outperforms the VRM-Vanilla which does not use textual preference data.
For instance, when using the LLaVA-1.5-7B model, RoVRM can outperform VRM-Vanilla by 8.4 points on the LLaVA-Bench.
We also observe this consistent phenomenon on the LLaVA-1.5-13B model.
Moreover, from the results, we find that RoVRM significantly reduces visual hallucinations, \textit{e.g.}, lowering the hallucination rate by 13.2 points in the LLaVA-1.5-7B model.
We attribute this improvement to the extensive use of textual preference data, which improves VRM's capacity to evaluate facticity.
Interestingly, we also find that RoVRM enables the LLaVA-1.5 models to outperform stronger LVLMs, with the LLaVA-1.5-7B model even surpassing the LLaVA-1.5-13B model on most of the benchmarks, such as MMHalBench and LLaVA-Bench.
This finding shows a promising direction for achieving \textit{weak-to-strong generalization} \cite{burns2023weak}.

\subsubsection{Results of Reinforcement Learning}
Compared to best-of-$n$ sampling, RL typically requires a more robust reward model: The reward model not only evaluates responses as “good” or “bad” but also provides an accuracy score margin between the responses \cite{zhou2024prior}.
From the results, we find that RoVRM fulfills this requirement more effectively than VRM-Vanilla, resulting in improved RL training performance in LVLMs. 
For instance, in RL training on the LLaVA-1.5-7B model, RoVRM surpasses VRM-Vanilla by 7.58 points on MM-Instruct.
This finding demonstrates that RoVRM is robust and can deliver high-quality reward signals across various alignment techniques.
Additionally, we observe that RL training reduces hallucinations but slightly decreases the “Cover.” metric, which is consistent with the findings of \citet{meng2024simpo}'s work and DPO training in Table \ref{tab:dpo_training}. 
We conjecture that preference alignment training may slightly hurt the instruction-following capability of LVLMs \cite{wang2024hybrid}.

Furthermore, compared to RoVRM-Random, RoVRM shows better performance across all benchmarks. 
This indicates that optimal transport-based preference data selection outperforms random selection. 
However, RoVRM-Random also significantly improves performance over VRM-Vanilla. 
We attribute this to the fact that RoVRM-Random also collects some textual preference data when training a VRM.

\begin{figure}
    \centering
    \scalebox{0.87}{
    \tikzstyle{every node}=[scale=1.0]
\hspace*{-0.2cm}
\begin{tikzpicture}
    \scriptsize{
      \begin{axis}[
        at={(-11.5em,8em)},
        ymajorgrids,
        grid style=dashed,
        ybar=2pt,
        enlarge x limits=0.5,
        xtick align=inside,
        height=.25\textwidth,
        width=.28\textwidth,
        bar width=1.0em,
        xlabel={MM-Instrut},
        ylabel={WinRate},
        symbolic x coords={{1}, {2}, {3}, {4}},
        xtick=data,
        nodes near coords align={vertical},
        xticklabels={{5k},{10k},{20k},{40k}},
        x tick label style={
             rotate=0,
             anchor=center,
             scale=1.2,
             xshift=-.0em,
             yshift=-0.8em
         },
        enlarge x limits=0.2,
        ylabel style={yshift=-1.3em,scale=1.3},
        xlabel style={yshift=0.8em,align=center,scale=1.2},
        ymin=44.0,
        ymax=61.0,
        ytick={40.0,45.0,...,60.0},
        yticklabel style={/pgf/number format/fixed,/pgf/number format/fixed zerofill,/pgf/number format/precision=1,rotate=0,scale=1.20},
        legend entries={\large Best-of-$n$ Sampling , \large Reinforcement Learning},
        legend columns=-1,
        legend style={
                minimum height=4ex,
                inner sep=.7pt,
                at={(-0.15, 1.15)},
                anchor=center,
                nodes={scale=0.8},
                legend cell align=center,
			    legend plot pos=left,
                draw=black!20,
                /tikz/every even column/.append style={column sep=0.20cm}},
        ]
        \addplot+[draw=red!40,fill=myred,area legend,error bars/.cd,y dir=both,y explicit,error bar style={color=myblack,line width=0.7pt}]
        table [
            x=X, y=Y, 
            y error plus=YerrPlus, y error minus=YerrMinus
        ] {
            X   Y   YerrPlus  YerrMinus
            1       52.74   0.65    0.65
            2       56.42   0.33    0.33
            3       58.94   0.57    0.57
            4       58.87   0.9     0.9
        };
        \addplot+[draw=blue!40,fill=myblue,area legend,error bars/.cd,y dir=both,y explicit,error bar style={color=myblack,line width=0.7pt}]
        table [
            x=X, y=Y, 
            y error plus=YerrPlus, y error minus=YerrMinus
        ] {
            X   Y   YerrPlus  YerrMinus
            1       47.13   0.49    0.49
            2       54.76   0.73    0.73
            3       56.32   0.82    0.82
            4       55.34   0.41    0.41
        };
      \end{axis}

      \begin{axis}[
        at={(-30em,8em)},
        ymajorgrids,
        grid style=dashed,
        ybar=2pt,
        enlarge x limits=0.5,
        xtick align=inside,
        height=.25\textwidth,
        width=.28\textwidth,
        bar width=1.0em,
        xlabel={LLaVA$^\mathrm{W}$},
        ylabel={Score},
        symbolic x coords={{1}, {2}, {3}, {4}},
        xtick=data,
        nodes near coords align={vertical},
        xticklabels={{5k},{10k},{20k},{40k}},
        x tick label style={
             rotate=0,
             anchor=center,
             scale=1.2,
             xshift=-.0em,
             yshift=-0.8em
         },
        enlarge x limits=0.2,
        ylabel style={yshift=-1.3em,scale=1.3},
        xlabel style={yshift=0.8em,align=center,scale=1.2},
        ytick={68.0,70.0,...,80.0},
        yticklabel style={/pgf/number format/fixed,/pgf/number format/fixed zerofill,/pgf/number format/precision=1,rotate=0,scale=1.20},
        ]
        \addplot+[draw=red!40,fill=myred,error bars/.cd,y dir=both,y explicit,error bar style={color=myblack,line width=0.7pt}]
        table [
            x=X, y=Y, 
            y error plus=YerrPlus, y error minus=YerrMinus
        ] {
            X   Y   YerrPlus  YerrMinus
            1       74.2    0.24    0.24
            2       75.6    0.16    0.16
            3       78.8    0.33    0.33
            4       78.1    0.41    0.41
        };
        
        \addplot+[draw=blue!40,fill=myblue,error bars/.cd,y dir=both,y explicit,error bar style={color=myblack,line width=0.7pt}]
        table [
            x=X, y=Y, 
            y error plus=YerrPlus, y error minus=YerrMinus
        ] {
            X   Y   YerrPlus  YerrMinus
            1       69.8    0.49    0.49
            2       74.2    0.41    0.41
            3       76.3    0.33    0.33
            4       72.2    0.33    0.33
        };
  
      \end{axis}  
    }
	\node [anchor=center]at (-32ex,9.2ex) {\scalebox{1.5}{(a) Textual Preference Data}};
	\scriptsize{
      \begin{axis}[
        at={(-11.5em,-9em)},
        ymajorgrids,
        grid style=dashed,
        legend style={at={(0.02,0.65)}, anchor=south west},
        legend cell align={left},
        ybar=2pt,
        enlarge x limits=0.5,
        xtick align=inside,
        height=.25\textwidth,
        width=.28\textwidth,
        bar width=1.0em,
        xlabel={MM-Instrut},
        ylabel={WinRate},
        symbolic x coords={{1}, {2}, {3}, {4}},
        xtick=data,
        nodes near coords align={vertical},
        xticklabels={{5k},{10k},{20k},{40k}},
        x tick label style={
             rotate=0,
             anchor=center,
             scale=1.2,
             xshift=-.0em,
             yshift=-0.8em
         },
        enlarge x limits=0.2,
        ylabel style={yshift=-1.3em,scale=1.3},
        xlabel style={yshift=0.8em,align=center,scale=1.2},
        ytick={50.0,52.0,...,62.0},
        yticklabel style={/pgf/number format/fixed,/pgf/number format/fixed zerofill,/pgf/number format/precision=1,rotate=0,scale=1.20},
        ]
        \addplot+[draw=red!40,fill=myred,error bars/.cd,y dir=both,y explicit,error bar style={color=myblack,line width=0.7pt}]
        table [
            x=X, y=Y, 
            y error plus=YerrPlus, y error minus=YerrMinus
        ] {
            X   Y   YerrPlus  YerrMinus
            1       57.32   0.49    0.49
            2       61.85   0.41    0.41
            3       58.94   0.57    0.57
            4       57.14   0.41    0.41
        };
        \addplot+[draw=blue!40,fill=myblue,error bars/.cd,y dir=both,y explicit,error bar style={color=myblack,line width=0.7pt}]
        table [
            x=X, y=Y, 
            y error plus=YerrPlus, y error minus=YerrMinus
        ] {
            X   Y   YerrPlus  YerrMinus
            1       54.18   0.24    0.24
            2       58.7    0.41    0.41
            3       56.32   0.41    0.41
            4       57.21   0.57    0.57
        };
      \end{axis}

      \begin{axis}[
        at={(-30em,-9em)},
        ymajorgrids,
        grid style=dashed,
        ybar=2pt,
        enlarge x limits=0.5,
        xtick align=inside,
        height=.25\textwidth,
        width=.28\textwidth,
        bar width=1.0em,
        xlabel={LLaVA$^\mathrm{W}$},
        ylabel={Score},
        symbolic x coords={{1}, {2}, {3}, {4}},
        xtick=data,
        nodes near coords align={vertical},
        xticklabels={{5k},{10k},{20k},{40k}},
        x tick label style={
             rotate=0,
             anchor=center,
             scale=1.2,
             xshift=-.0em,
             yshift=-0.8em
         },
        enlarge x limits=0.2,
        ylabel style={yshift=-1.3em,scale=1.3},
        xlabel style={yshift=0.8em,align=center,scale=1.2},
        ymax=83,
        ymin=69,
        ytick={70.0,72.0,...,82.0},
        yticklabel style={/pgf/number format/fixed,/pgf/number format/fixed zerofill,/pgf/number format/precision=1,rotate=0,scale=1.20},
        ]
        \addplot+[draw=red!40,fill=myred,error bars/.cd,y dir=both,y explicit,error bar style={color=myblack,line width=0.7pt}]
        table [
            x=X, y=Y, 
            y error plus=YerrPlus, y error minus=YerrMinus
        ] {
            X   Y   YerrPlus  YerrMinus
            1       78.8    0.49    0.49
            2       82.0    0.24    0.24
            3       78.9    0.33    0.33
            4       79.3    0.33    0.33
        };
        \addplot+[draw=blue!40,fill=myblue,error bars/.cd,y dir=both,y explicit,error bar style={color=myblack,line width=0.7pt}]
        table [
            x=X, y=Y, 
            y error plus=YerrPlus, y error minus=YerrMinus
        ] {
            X   Y   YerrPlus  YerrMinus
            1       74.3    0.24    0.24
            2       78.0    0.57    0.57
            3       76.3    0.33    0.33
            4       75.1    0.33    0.33
        };        
      \end{axis}
 }
    \node [anchor=center]at (-32ex,-29ex) {\scalebox{1.5}{(b) Image Caption-based Preference Data}};
\end{tikzpicture}}
    \vspace{-2mm}
    \caption{
    We train RoVRM with varying amounts of textual and image caption preference data. Experiments are conducted on the LLaVA-1.5-7B model using three different seeds, and we report the average results along with their standard deviation.
    }
    \label{fig:diff_selection_sizes}
\end{figure}
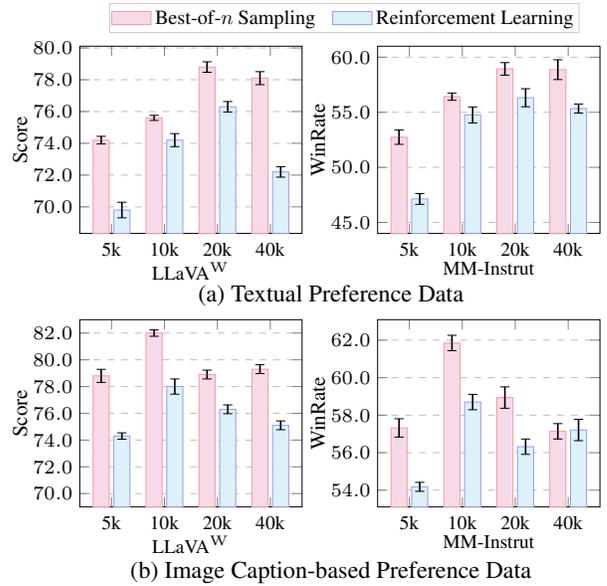

\subsection{Ablation Study}
We present detailed ablation studies to investigate the effects of three-phase progressive training and our preference data selection approach. 
The experiments are conducted on the LLaVA-1.5-7B model and the impacts of removing each approach were thoroughly examined. 
Furthermore, we study the impact of eliminating the distinct designs of phases one and two.
The results are summarized in Table \ref{tab:ablation_study}.
Through the results, we can see that three-phase progressive training significantly improves the performance of RoVRM in both best-of-$n$ sampling and RL.
Notably, removing phase one leads to a substantial performance decline (\textit{e.g.}, a loss of 10.7 points on the LLaVA-Bench for best-of-$n$ sampling), highlighting the importance of textual preference data in training RoVRM. 
Likewise, removing image caption-based preference data also results in performance loss, indicating the need to address the task gap.
Additionally, we see that using the preference data selection can train a better RoVRM.
It shows the effectiveness of using optimal transport to conduct preference data selection.

\begin{figure}[t]
    \centering
    \scalebox{0.86}{
    \begin{tikzpicture}
\hspace*{-0.1cm}
\scriptsize{
\begin{axis}
[
    at={(0,0)},
    ymajorgrids=true,
    xmajorgrids=true,
    clip=false,
    grid style=dashed,
    height=.25\textwidth,
    width=.28\textwidth,
    legend columns=3,
    legend style={
        anchor=south east,
        legend cell align={right},
        at={(4.05cm, 1.2)},
        anchor=center,
        font=\scriptsize,
        draw=gray,
        cells={anchor=west},
        fill opacity=0.7
    },
    ylabel={\scriptsize{HalRate ($\downarrow$)}},
    ylabel style={yshift=-2.0ex,xshift=0em,scale=1.4},
    xlabel={\scriptsize{Training Step}},
    xlabel style={xshift=0em,yshift=0.2em,scale=1.4},
    xmin=0,
    xmax=40,
    ymax=61,
    ymin=44,
    xtick={0,10,20,30,40},
    xticklabels={100,300,500,700,900},  
    ytick={45,50,55,60},
    x tick label style={/pgf/number format/fixed, /pgf/number format/fixed zerofill, /pgf/number format/precision=1,scale=1.4},
    y tick label style={/pgf/number format/fixed,/pgf/number format/fixed zerofill,/pgf/number format/precision=1, scale=1.4},
]
\addplot[
    color=myblack,
    line width=1pt
] coordinates {
    (0,55.0) (40,55.0)
};
\addlegendentry{\scalebox{1.1}{LLaVA-1.5-7B}};

\addplot+[
    error_red,
    mark=none,
    mark size=0.8pt,
    line width=1.0pt,
    mark options={fill=error_red,draw=error_red,line width=1.5pt},
    error bars/.cd,y dir=both,y explicit,error bar style={color=error_red,line width=1.0pt,mark size=1.5pt}
]
    table [
        x=X, y=Y, 
        y error plus=YerrPlus, y error minus=YerrMinus
    ] {
        X   Y   YerrPlus  YerrMinus
        0       54.9    0.24    0.24
        5       52.1    0.82    0.82
        10      51.0    0.9     0.9 
        15      50.3    0.57    0.57
        20      49.7    1.06    1.06
        25      47.9    0.9     0.9 
        30      48.3    0.57    0.57
        35      47.6    0.44    0.44
        40      47.9    0.9     0.9
    };

\addlegendentry{\scalebox{1.1}{RL w/ RoVRM}}

\addplot+[
    error_blue,
    mark=none,
    mark size=0.8pt,
    line width=1.0pt,
    mark options={fill=error_blue,draw=error_blue,line width=1.5pt},
    error bars/.cd,y dir=both,y explicit,error bar style={color=error_blue,line width=1.0pt,mark size=1.5pt}
    ]
    table [
        x=X, y=Y, 
        y error plus=YerrPlus, y error minus=YerrMinus
    ] {
        X   Y   YerrPlus  YerrMinus
        0       55.6    0.57    0.57
        5       53.5    0.57    0.57
        10      52.8    1.14    1.14
        15      52.1    1.71    1.71
        20      54.9    1.14    1.14
        25      54.2    1.71    1.71
        30      56.3    1.63    1.63
        35      52.7    1.22    1.22
        40      52.5    0.98    0.98
    };
\addlegendentry{\scalebox{1.1}{RL w/ VRM-Vanilla}};

\end{axis}

\begin{axis}
[
    at={(4.7cm,0)},
    ymajorgrids=true,
    xmajorgrids=true,
    clip=false,
    grid style=dashed,
    height=.25\textwidth,
    width=.28\textwidth,
    legend style={
        at={(0.95,0.06)},
        anchor=south east,
        legend cell align={right},
        font=\scriptsize,
        draw=gray,
        cells={anchor=west},
        fill opacity=0.7
    },
    ylabel={\scriptsize{Score ($\uparrow$)}},
    ylabel style={yshift=-2.0ex,xshift=0em,scale=1.4},
    xlabel={\scriptsize{Training Step}},
    xlabel style={xshift=0em,yshift=0.2em,scale=1.4},
    xmin=0,
    xmax=40,
    ymax=81,
    ymin=64,
    xtick={0,10,20,30,40},
    xticklabels={100,300,500,700,900},  
    ytick={65,70,75,...,80},
    x tick label style={/pgf/number format/fixed, /pgf/number format/fixed zerofill, /pgf/number format/precision=1,scale=1.4},
    y tick label style={/pgf/number format/fixed,/pgf/number format/fixed zerofill,/pgf/number format/precision=1, scale=1.4},
]
\addplot[
    color=myblack,
    line width=1pt
] coordinates {
    (0,66.7) (40,66.7)
};

\addplot+[
    error_red,
    mark=none,
    mark size=0.8pt,
    line width=1.0pt,
    mark options={fill=error_red,draw=error_red,line width=1.5pt},
    error bars/.cd,y dir=both,y explicit,error bar style={color=error_red,line width=1.0pt,mark size=1.5pt}
]
    table [
        x=X, y=Y, 
        y error plus=YerrPlus, y error minus=YerrMinus
    ] {
        X   Y   YerrPlus  YerrMinus
        0       69.0    0.82    0.82
        5       73.1    0.9     0.9
        10      75.5    1.14    1.14
        15      75.8    0.57    0.57
        20      76.5    0.73    0.73
        25      76.7    0.73    0.73
        30      77.3    0.57    0.57
        35      78.0    0.73    0.73
        40      78.2    0.9     0.9
    };
\addplot+[
    error_blue,
    mark=none,
    mark size=0.8pt,
    line width=1.0pt,
    mark options={fill=error_blue,draw=error_blue,line width=1.5pt},
    error bars/.cd,y dir=both,y explicit,error bar style={color=error_blue,line width=1.0pt,mark size=1.5pt}
    ]
    table [
        x=X, y=Y, 
        y error plus=YerrPlus, y error minus=YerrMinus
    ] {
        X   Y   YerrPlus  YerrMinus
        0       67.8    1.14    1.14
        5       72.5    1.14    1.14
        10      72.8    1.31    1.31
        15      73.3    0.82    0.82
        20      73.2    1.39    1.39
        25      73.6    0.73    0.73
        30      72.4    1.14    1.14
        35      71.5    0.82    0.82
        40      71.1    0.65    0.65
    };
\end{axis}
}

\end{tikzpicture}}
    \vspace{-6mm}
    \caption{
    Performance during RL training is evaluated on the  MMHalBench (left) and LLaVA-Bench (right) benchmarks using three different seeds.
    }
    \vspace{-6mm}
    \label{fig:rl_training}
\end{figure}
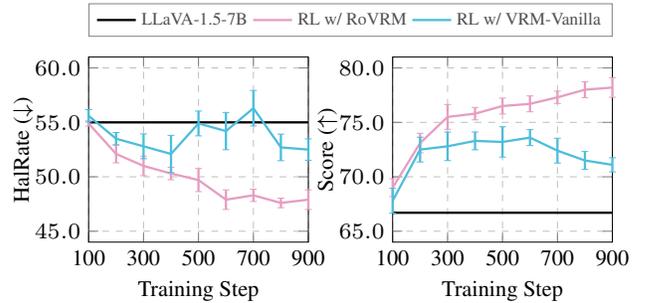

\subsection{Analysis}

\subsubsection{Performance on Different Numbers of Selected Preference Samples}
We investigate the impact of different numbers of selected preference samples using a three-phase progressive training with LLaVA-Bench and MM-Instruct.
We test sample sizes of 5k, 10k, 20k, and 40k, alongside 20k image caption-based preference samples (Figure \ref{fig:diff_selection_sizes}(a)). 
Our results show that using 20k textual preference samples yields strong performance, even outperforming the 40k sample scenario. 
Consequently, we choose 20k textual preference samples for phase one to train our RoVRM.
Similarly, we evaluate sample sizes of 5k, 10k, 20k, and 40k for phase two, \textit{i.e.,} image caption-based preference data selection (Figure \ref{fig:diff_selection_sizes}(b)), identifying 10k as the optimal sample size.

\begin{figure}[t]
    \centering
    \scalebox{0.85}{
    \begin{tikzpicture}
\hspace*{-0.00cm}
\scriptsize{
\begin{axis}
[
    at={(0,0)},
    ymajorgrids=true,
    xmajorgrids=true,
    clip=false,
    grid style=dashed,
    height=.25\textwidth,
    width=.28\textwidth,
    legend columns=2,
    legend style={
        anchor=south east,
        legend cell align={right},
        at={(4.05cm, 1.2)},
        anchor=center,
        font=\scriptsize,
        draw=gray,
        cells={anchor=west},
        fill opacity=0.7
    },
    ylabel={\scriptsize{HalRate ($\downarrow$)}},
    ylabel style={yshift=-2.0ex,xshift=0em,scale=1.4},
    xlabel={\scriptsize{Number of VPD}},
    xlabel style={xshift=0em,yshift=0.3em,scale=1.4},
    xmin=0,
    xmax=40,
    ymax=61,
    ymin=39,
    xtick={0,10,20,30,40},
    xticklabels={0k,10k,20k,30k,40k},  
    ytick={40,45,50,55,60},
    x tick label style={/pgf/number format/fixed, /pgf/number format/fixed zerofill, /pgf/number format/precision=1,scale=1.4},
    y tick label style={/pgf/number format/fixed,/pgf/number format/fixed zerofill,/pgf/number format/precision=1, scale=1.4},
]
\addplot[
    color=myblack,
    line width=1.0pt,,
    dash pattern=on 2pt off 2pt
] coordinates {
    (0,47.9) (40,47.9)
};
\addlegendentry{\scalebox{1.2}{BoS w/ VRM-Vanilla}}
\addplot[
    color=myblack,
    line width=1.0pt,
] coordinates {
    (0,53.2) (40,53.2)
};
\addlegendentry{\scalebox{1.2}{RL w/ VRM-Vanilla}};

\addplot+[
    error_red,
    mark=none,
    line width=1.0pt,
    dash pattern=on 2pt off 2pt,
    error bars/.cd,y dir=both,y explicit,error bar style={color=error_red,line width=1.0pt,,mark size=1.5pt}
]
    table [
        x=X, y=Y, 
        y error plus=YerrPlus, y error minus=YerrMinus
    ] {
        X   Y   YerrPlus  YerrMinus
        0       47.6    0.33    0.33
        2       46.5    0.64    0.64
        5       46.1    0.57    0.57
        10      44.5    0.44    0.44
        20      43.4    0.33    0.33
        30      43.3    1.22    1.22
        40      43.3    0.41    0.41
    };
\addlegendentry{\scalebox{1.2}{BoS w/ RoVRM}};

\addplot+[
    error_blue,
    mark=none,
    mark size=0.8pt,
    line width=1.0pt,
    mark options={fill=error_blue,draw=error_blue,line width=1.5pt},
    error bars/.cd,y dir=both,y explicit,error bar style={color=error_blue,line width=1.0pt,,mark size=1.5pt}
    ] 
    table [
        x=X, y=Y, 
        y error plus=YerrPlus, y error minus=YerrMinus
    ] {
        X   Y   YerrPlus  YerrMinus
        0       58.0    1.14    1.14
        2       56.2    0.73    0.73
        5       53.4    0.57    0.57
        10      52.8    1.06    1.06
        20      49.3    0.57    0.57
        30      48.6    0.33    0.33
        40      48.3    0.57    0.57
    };
\addlegendentry{\scalebox{1.2}{RL w/ RoVRM}}

\end{axis}

\begin{axis}
[
    at={(4.7cm,0)},
    ymajorgrids=true,
    xmajorgrids=true,
    clip=false,
    grid style=dashed,
    height=.25\textwidth,
    width=.28\textwidth,
    legend style={
        at={(0.95,0.06)},
        anchor=south east,
        legend cell align={right},
        font=\scriptsize,
        draw=gray,
        cells={anchor=west},
        fill opacity=0.7
    },
    ylabel={\scriptsize{Score ($\uparrow$)}},
    ylabel style={yshift=-2.0ex,xshift=0em,scale=1.4},
    xlabel={\scriptsize{Number of VPD}},
    xlabel style={xshift=0em,yshift=0.3em,scale=1.4},
    xmin=0,
    xmax=40,
    ymax=86,
    ymin=64,
    xtick={0,10,20,30,40},
    xticklabels={0k,10k,20k,30k,40k},  
    ytick={65,70,75,...,85},
    x tick label style={/pgf/number format/fixed, /pgf/number format/fixed zerofill, /pgf/number format/precision=1,scale=1.4},
    y tick label style={/pgf/number format/fixed,/pgf/number format/fixed zerofill,/pgf/number format/precision=1, scale=1.4},
]
\addplot[
    color=myblack,
    line width=1.0pt,
    dash pattern=on 2pt off 2pt
] coordinates {
    (0,73.6) (40,73.6)
};
\addplot[
    color=myblack,
    line width=1.0pt,
] coordinates {
    (0,72.8) (40,72.8)
};

\addplot+[
    error_red,
    mark=none,
    mark size=0.8pt,
    line width=1.0pt,
    dash pattern=on 2pt off 2pt,
    mark options={fill=error_red,draw=error_red,line width=1.5pt},
    error bars/.cd,y dir=both,y explicit,error bar style={color=error_red,line width=1.0pt,,mark size=1.5pt}
]
    table [
        x=X, y=Y, 
        y error plus=YerrPlus, y error minus=YerrMinus
    ] {
        X   Y   YerrPlus  YerrMinus
        0       71.6    1.8     1.8
        2       74.9    0.73    0.73
        5       77.1    1.14    1.14
        10      79.9    1.31    1.31
        20      80.4    0.82    0.82
        30      81.4    0.9     0.9
        40      80.9    0.82    0.82
    };

\addplot+[
    error_blue,
    mark=none,
    mark size=0.8pt,
    line width=1.0pt,
    mark options={fill=error_blue,draw=error_blue,line width=1.5pt},
    error bars/.cd,y dir=both,y explicit,error bar style={color=error_blue,line width=1.0pt,,mark size=1.5pt}
    ]
    table [
        x=X, y=Y, 
        y error plus=YerrPlus, y error minus=YerrMinus
    ] {
        X   Y   YerrPlus  YerrMinus
        0       73.1    0.24    0.24
        2       74.1    0.33    0.33
        5       73.8    0.9     0.9
        10      73.6    1.22    1.22
        20      74.3    0.9     0.9
        30      74.8    1.22    1.22
        40      75.3    1.31    1.31
    };
\end{axis}
}

\end{tikzpicture}}
    \vspace{-6mm}
    \caption{
    Performance of best-of-$n$ sampling (BoS) and RL on MMHalBench (left) and LLaVA-Bench (right) across three different seeds. The RoVRM model is trained with varying amounts of visual preference data (VPD): 0k, 1k, 5k, 10k, 20k, 30k, and 40k.
    }
    \label{fig:few_shot}
\end{figure}
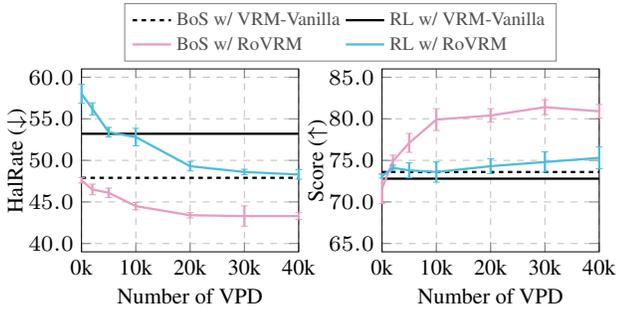

\begin{table}
    \centering
    \vspace{-1mm}
    \scalebox{0.80}{
    \begin{tabular}{lcccc}
\toprule[1.1pt]
\multirow{2}{*}{Method}  
& \multicolumn{2}{c}{AMBER}                  
& LLaVA$^\mathrm{{W}}$
& MMIns
\\ \cmidrule(l){2-3} \cmidrule(l){4-4} \cmidrule(l){5-5}
&  Cover. $\uparrow$ & HalRate $\downarrow$ & Score $\uparrow$  & WinRate $\uparrow$  \\ \midrule
LLaVA-1.5-7B  &50.3  &37.1 &66.7 &46.16          \\ \midrule
\ \ \ \ \ +DPO &49.6  &22.2  &80.9  & 56.09 \\ \rowcolor{gray!20}
\ \ \ \ \ +RoDPO &\bf{50.7}  &\bf{17.6}  &\bf{83.7}  & \bf{73.91} \\ \midrule
LLaVA-1.5-13B  &\bf{50.6} &37.2 &75.6 & 50.00          \\ \midrule
\ \ \ \ \ +DPO &49.2  &15.7  &84.2  & 65.63 \\
\rowcolor{gray!20}
\ \ \ \ \ +RoDPO &49.8  &\bf{12.8}  &\bf{86.4}  & \bf{78.72} \\ 
\bottomrule[1.1pt]
\end{tabular}}
    \caption{Performance on the direct preference optimization.}
    \vspace{-5mm}
    \label{tab:dpo_training}
\end{table}

\subsubsection{Comparison of RL Training Process on Different VRMs}
Figure \ref{fig:rl_training} illustrates the performance of the LLaVA-1.5-7B model comparing RL training with VRM-Vanilla and RoVRM. 
The results show that RL training with RoVRM improves performance more effectively than VRM-Vanilla.
Additionally, we observe that RoVRM can lead to a more stable RL training process by mitigating reward over-optimization \cite{gao2023scaling}.

\begin{figure}[t]
    \centering
    \scalebox{0.86}{\tikzstyle{every node}=[scale=1.0]
\hspace*{-0.2cm}
\begin{tikzpicture}
    \scriptsize{
      \begin{axis}[
        at={(-11.5em,8em)},
        ymajorgrids,
        grid style=dashed,
        ybar=2pt,
        enlarge x limits=0.5,
        xtick align=inside,
        height=.25\textwidth,
        width=.28\textwidth,
        bar width=1.0em,
        xlabel={Sampling Size},
        ylabel={Score ($\uparrow$)},
        symbolic x coords={{1}, {2}, {3}, {4}},
        xtick=data,
        nodes near coords align={vertical},
        xticklabels={{4},{8},{16},{32}},
        x tick label style={
             rotate=0,
             anchor=center,
             scale=1.3,
             xshift=-.0em,
             yshift=-0.8em
         },
        enlarge x limits=0.2,
        ylabel style={yshift=-1.3em,scale=1.3},
        xlabel style={yshift=0.8em,align=center,scale=1.3},
        ymin=68,
        ymax=88,
        ytick={70,75,...,85.0},
        yticklabel style={/pgf/number format/fixed,/pgf/number format/fixed zerofill,/pgf/number format/precision=1,rotate=0,scale=1.20},
        legend entries={\scalebox{1.4}{BoS w/ RoVRM}, \scalebox{1.4}{BoS w/ VRM-Vanilla}},
        legend columns=-1,
        legend style={
                minimum height=4ex,
                nodes={scale=0.85, transform shape},
                inner sep=.7pt,
                at={(-0.18, 1.15)},
                anchor=center,
                legend cell align=center,
			    legend plot pos=left,
                draw=black!20,
                /tikz/every even column/.append style={column sep=0.20cm}},
        ]
        \addplot+[draw=red!40,fill=myred,area legend,error bars/.cd,y dir=both,y explicit,error bar style={color=myblack,line width=0.7pt}]
        table [
            x=X, y=Y, 
            y error plus=YerrPlus, y error minus=YerrMinus
        ] {
            X   Y   YerrPlus  YerrMinus
            1       78.6    1.14    1.14
            2       82.3    0.73    0.73
            3       83.1    0.73    0.73
            4       84.2    0.9     0.9
        };
        \addplot+[draw=blue!40,fill=myblue,area legend,error bars/.cd,y dir=both,y explicit,error bar style={color=myblack,line width=0.7pt}]
        table [
            x=X, y=Y, 
            y error plus=YerrPlus, y error minus=YerrMinus
        ] {
            X   Y   YerrPlus  YerrMinus
            1       72.5    1.06    1.06
            2       73.7    0.41    0.41
            3       74.8    0.24    0.24
            4       77.5    0.82    0.82
        };
      \end{axis}

      \begin{axis}[
        at={(-30em,8em)},
        ymajorgrids,
        grid style=dashed,
        ybar=2pt,
        enlarge x limits=0.5,
        xtick align=inside,
        height=.25\textwidth,
        width=.28\textwidth,
        bar width=1.0em,
        xlabel={Sampling Size},
        ylabel={HalRate ($\downarrow$)},
        symbolic x coords={{1}, {2}, {3}, {4}},
        xtick=data,
        nodes near coords align={vertical},
        xticklabels={{4},{8},{16},{32}},
        x tick label style={
             rotate=0,
             anchor=center,
             scale=1.3,
             xshift=-.0em,
             yshift=-0.8em
         },
        enlarge x limits=0.2,
        ylabel style={yshift=-1.3em,scale=1.3},
        xlabel style={yshift=0.8em,align=center,scale=1.3},
        ymax=61,
        ymin=29,
        ytick={30.0,35.0,...,60.0},
        yticklabel style={/pgf/number format/fixed,/pgf/number format/fixed zerofill,/pgf/number format/precision=1,rotate=0,scale=1.20},
        ]
        \addplot+[draw=red!40,fill=myred,error bars/.cd,y dir=both,y explicit,error bar style={color=myblack,line width=0.7pt}]
        table [
            x=X, y=Y, 
            y error plus=YerrPlus, y error minus=YerrMinus
        ] {
            X   Y   YerrPlus  YerrMinus
            1       52.4    2.29    2.29
            2       40.6    0.9     0.9
            3       39.6    0.82    0.82
            4       37.5    0.82    0.82
        };
        
        \addplot+[draw=blue!40,fill=myblue,error bars/.cd,y dir=both,y explicit,error bar style={color=myblack,line width=0.7pt}]
        table [
            x=X, y=Y, 
            y error plus=YerrPlus, y error minus=YerrMinus
        ] {
            X   Y   YerrPlus  YerrMinus
            1       58.3    1.71    1.71
            2       48.6    1.14    1.14
            3       45.5    0.24    0.24
            4       43.4    0.33    0.33
        };
  
      \end{axis}  
    }
\end{tikzpicture}}
    \vspace{-3mm}
    \caption{
    Performance of best-of-$n$ sampling (BoS) with different sampling sizes: 4, 8, 16, and 32.
    }
    \vspace{-0mm}
    \label{fig:diff_sampling_sizes_BoS}
\end{figure}
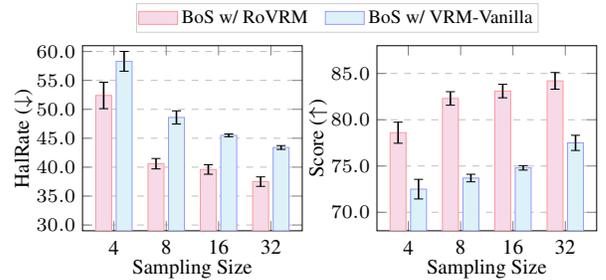

\begin{table}[t]
    \centering
    \scalebox{0.76}{
    \begin{tabular}{lcccc}
\toprule[1.1pt]
\multirow{2}{*}{Evaluation Result} & \multicolumn{2}{c}{Best-of-$n$ Sampling} & \multicolumn{2}{c}{Reinforcement Learning} \\  \cmidrule(l){2-3} \cmidrule(l){4-5}
& LLaVA$^\mathrm{W}$               & MMIns              & LLaVA$^\mathrm{W}$      & \multicolumn{1}{c}{MMIns}     \\ \midrule
RoVRM is better  &27   &33    &25    &44   \\
VRM-Vanilla is better  &12   &23    &15    &23   \\
Tie     &21   &43    &20    &32   \\ \midrule
P-value     & \multicolumn{2}{c}{0.0133} & \multicolumn{2}{c}{0.0035}            \\
\bottomrule[1.1pt]
\end{tabular}}
    \caption{
    The results of human evaluation.
    We report the statistical significance of the differences between the RoVRM and VRM-Vanilla models using the McNemar test.
    }
    \vspace{-5mm}
    \label{tab:human_evaluation}
\end{table}

\subsubsection{Enabling Few-Shot Learning in VRM}
Figure \ref{fig:few_shot} shows RoVRM's performance with different numbers of visual preference data. Note that when the visual preference dataset is small (\textit{i.e.}, 1k, 5k, and 10k), we use the entire dataset without image caption-based preference data selection.
From the results, we find that pre-training with textual preference data enables effective few-shot learning in VRM \cite{wang2020generalizing}.
Based on these textual preferences, the reward model quickly generalizes to vision-language tasks using only a few visual preference samples.
Notably, using only 5k visual preference samples can achieve a performance comparable to that of VRM-Vanilla trained with 83k samples.
However, while it is feasible to directly use a textual reward model (\textit{i.e.}, using 0k visual preference data) to optimize LVLM, the results are worse, particularly during RL training.

\subsubsection{Integration with Direct Preference Optimisation}
Despite bypassing reward model training, direct preference optimization (DPO) still requires preference data to train the language model with a ranking-loss function. 
Consequently, DPO also faces the challenge of limited visual preference data in LVLMs. 
To address this, we propose a \textbf{Ro}bust \textbf{DPO} (namely RoDPO) by integrating our three-phase progressive training and preference data selection.
Our experiments on the LLaVA-1.5-7B and -13B models show that RoDPO performs better than DPO, as summarized in Table \ref{tab:dpo_training}.

\subsubsection{Performance on Different Sampling Sizes}
We evaluate the performance of best-of-$n$ sampling with varying sample sizes using the LLaVA-1.5-7B model. 
Figure \ref{fig:diff_sampling_sizes_BoS} presents a comparison of RoVRM and VRM-Vanilla on the MMHalBench (left) and LLaVA-Bench (right) benchmarks. 
The experimental results indicate that RoVRM consistently enhances performance across different sampling sizes, highlighting its improved robustness.

\subsubsection{Human Evaluation}
In addition to automatic evaluation, we have conducted a comparative human evaluation of RoVRM and VRM-Vanilla on the LLaVA-1.5-7B model. 
Two independent evaluators, who remained anonymous, assessed the responses generated by these models for the same set of image-question pairs.
The results of this evaluation are presented in Table \ref{tab:human_evaluation}. 
These findings align with the outcomes of the automatic evaluation, demonstrating that our RoVRM consistently outperforms VRM-Vanilla. The statistically significant differences observed between the two models further underscore the distinct advantages of RoVRM.

See more analysis in \textbf{Appendix B}.

\section{Conclusion}
In this paper, we focus on improving the human-preference alignment for LVLMs.
We present a \textbf{Ro}bust \textbf{V}isual \textbf{R}eward \textbf{M}odel (namely RoVRM) via three-phase progressive training and preference data selection approaches.
Our extensive experiments demonstrate that our RoVRM significantly outperforms the traditional visual reward model.

\section{Acknowledgments}
This work was supported in part by the National Science Foundation of China (Nos. 62276056 and U24A20334), the Natural Science Foundation of Liaoning Province of China (2022-KF-26-01), the Fundamental Research Funds for the Central Universities (Nos. N2216016 and N2316002), the Yunnan Fundamental Research Projects (No. 202401BC070021), and the Program of Introducing Talents of Discipline to Universities, Plan 111 (No.B16009).

\bibliography{aaai25}

\appendix

\clearpage

\section{Appendix A: Experimental Details}
\label{sec:experimentalDetails}
\subsection{Settings}

\subsubsection{Best-of-$n$ Sampling}
we employed top-$p$ sampling with $p = 0.95$ and a temperature of 0.2 to generate eight candidate responses. 
We then picked the response with the highest reward score as the final output.

\subsubsection{RL Training}
We trained the LVLM using PPO via the \texttt{trlX} implementation\footnote{\url{https://github.com/CarperAI/trlx}}. 
The learning rates were set at 1e-5 for the language model and 5e-6 for the value model. 
Each PPO step used a batch size of 32, with 15k gradient steps and two mini-batch update epochs.
To address over-optimization, as noted by \citet{gao2023scaling}, we saved checkpoints at regular intervals during training. 
Specifically, we evaluated checkpoints every 100 steps using our validation set and selected the one with the highest reward score. 
The validation set, randomly sampled from LLaVA-Instruct-150K, consisted of 1k samples.
Following \citet{zhou2024prior} and \citet{wang2024hybrid}, we also employed a cold-start trick for PPO, to alleviate the damage caused by the inaccurate estimation of the early value model.
Specifically, we only updated the value model and did not update the policy model during the first 30 steps of RL training.
Additionally, following \citet{wang2024esrl}'s work, we standardized our reward scores using a reward queue, which stored the previous 1k reward scores to calculate the mean and variance.
All of our experiments were done on eight A800 GPUs.

\subsubsection{Preference Data Selection}
We utilized LoRA to pre-train the reward model, as described by \citet{xia2024less}. 
For each preference sample, we then extracted 8192-dimensional gradient features from the pre-trained model.

\subsubsection{DPO Training}
We used a batch size of 64, a learning rate of 1e-6, and trained for one epoch in DPO training. 
Except for these parameters, our setup matched \citet{rafailov2024direct}. 
In three-phase progressive training, the learning rate was 1e-6 in phase one and 5e-7 in phases two and three.

\subsubsection{Random Projection}
We employed random projection to reduce the dimensionality of gradient features, illustrated here with the example of computing the gradient feature for a preference sample $s_i$. 
Initially, we concatenated all gradients from $\nabla \mathcal{L}_{reward}(s_{i}; \theta_{warmup})$ to form the gradient feature $g^{\mathrm{init}}_{i}\in \mathbb{R}^{1\times d}$, where $d$ represents the number of trainable neurons. 
Although the LoRA training approach was employed to reduce $d$, it remained large.
Therefore, random projection was applied to further reduce the dimensionality.
Specifically, a random matrix $\mathrm{R}$ was first generated with dimensions $d\times k$, where $k$ denotes the target dimensionality, significantly lower than $d$. 
The elements of $\mathrm{R}$ were typically drawn from a standard normal distribution $\mathcal{N}(0,1)$.
Then, the high-dimensional $g^{\mathrm{init}}_{i}$ was projected into a lower-dimensional space by computing the matrix product:
\begin{eqnarray}
    g_{i} = g^{\mathrm{init}}_{i}\times \mathrm{R}
\end{eqnarray} 
According to the Johnson-Lindenstrauss Lemma \cite{xie2017survey}, random projection approximately preserved the pairwise Euclidean distances between our gradient features with high probability.

\subsection{Evaluation}
In this section, we describe how we compute the WinRate for the MM-Instruct benchmark.
Given the pairwise test responses $\{(x^{0}, r_a^{0}, r_b^{0}), \cdots, (x^{T}, r_a^{T}, r_b^{T})\}$, where $T$ is the number of the test set, we employ GPT-4 to annotate the preference of each pairwise response, including $P_a$, $P_b$, and $Tie$.
Here, $P_a$ denotes response $r_{a}$ is better than response $r_{b}$, $P_b$ denotes response $r_{b}$ is worse than response $r_{b}$, while $Tie$ denotes a tie between response $r_{a}$ and response $r_{b}$.
To address potential location bias in the evaluation \cite{gao2024llm}, we conduct two separate evaluations for each pair, alternating the order of $r_a$ and $r_b$. 
The final result is based on the evaluations where the preferences align consistently.
We can compute the WinRate for the models generating responses $r_a$ and $r_b$ based on these annotated preferences:
\begin{eqnarray}
    S_{\mathrm{WinRate}}^{a}=\frac{\mathrm{Count}(P_{a})}{T-\mathrm{Count}(Tie)} \\
    S_{\mathrm{WinRate}}^{b}=\frac{\mathrm{Count}(P_{b})}{T-\mathrm{Count}(Tie)}
\end{eqnarray}
where $\mathrm{Count}(\cdot)$ represents the number of occurrences of the specified preference.

\subsection{Image Caption Annotation}
For image caption annotation, we employed a template similar to that proposed by \citet{liu2024visual}. 
However, this approach raises two concerns: the potential unreliability of the annotations and associated ethical risks.
First, \texttt{GPT-4o} is widely regarded as a highly effective visual model, with its smaller variant, \texttt{GPT-4o-mini}, demonstrating comparable performance on image caption tasks across multiple publicly available benchmarks, such as Object HalBench and AMBER (see Table \ref{tab:main_experiments}). 
Furthermore, \texttt{GPT-4o-mini} is significantly more cost-effective. 
Given these factors, we consider \texttt{GPT-4o-mini} to be reliable and select it for our annotation process.
Second, ethical and moral risks are not a concern since the images are derived from open datasets, and GPT-4o functions as an open multimodal LLM.

We employed \texttt{Vision-LLM-Alignment}\footnote{\url{https://github.com/wangclnlp/Vision-LLM-Alignment}} to train our RoVRM, and performed best-of-$n$ sampling, RL training, and DPO training.

\section{Appendix B: More Analysis}
\subsubsection{Comparison with Other Preference Data Selection Approaches}
To validate the effectiveness of using optimal transport in our preference data approach, we compare it with other distance computation approaches. 
Specifically, we examine alternative ways of calculating the distance score in Eq. \ref{eq:distance_score}, such as \textbf{Cosine Similarity}, Kullback-Leibler divergence (\textbf{KL}), and \textbf{$\bm{L_{2}}$-norm}, instead of the optimal transport function $\mathrm{OT}(\cdot)$. 
The results are summarized in Table \ref{tab:diff_data_selection_approaches}.
From the results, we can observe that optimal transport significantly outperforms other approaches in preference data selection. 
This superiority underscores its effectiveness in identifying relevant textual preferences for training RoVRM, as optimal transport more accurately captures the transfer distance of preferences, whereas other approaches primarily measure similarity.

\begin{table}[]
    \centering
    \scalebox{0.75}{
    \begin{tabular}{lcccc}
\toprule[1.1pt]
& \multicolumn{2}{c}{Best-of-$n$ Sampling} & \multicolumn{2}{c}{Reinforcement Learning}   \\ \cmidrule(l){2-3} \cmidrule(l){4-5}
\multirow{-2}{*}{Method} & {\begin{tabular}[c]{@{}c@{}}AMBER\\ (HalRate $\downarrow$)\end{tabular}} & {\begin{tabular}[c]{@{}c@{}}LLaVA$^\mathrm{W}$\\ (Score $\uparrow$)\end{tabular}}  & {\begin{tabular}[c]{@{}c@{}}AMBER\\ (HalRate $\downarrow$)\end{tabular}} & {\begin{tabular}[c]{@{}c@{}}LLaVA$^\mathrm{W}$\\ (Score $\uparrow$)\end{tabular}}    \\ \midrule

VRM-Vanilla       &29.0 &73.6   &29.1 &72.8     \\ \midrule
RoVRM \\
+ Cosine Similarity &25.7    &78.9    &25.2   &75.1      \\
+ KL                &26.4    &75.4    &28.5   &72.2      \\
+ $\bm{L}_{2}$-norm &27.9    &76.6    &26.0   &74.3      \\
\rowcolor{gray!20}
+ Optimal Transport &\bf{23.9} &\bf{82.0} &\bf{23.4} &\bf{78.3} \\

\bottomrule[1.1pt]
\end{tabular}}
    \caption{
    Performance on the best-of-$n$ sampling and RL training with reward models trained via various preference data selection approaches.
    }
    \label{tab:diff_data_selection_approaches}
\end{table}

\subsubsection{Performance on Different Temperature Settings}
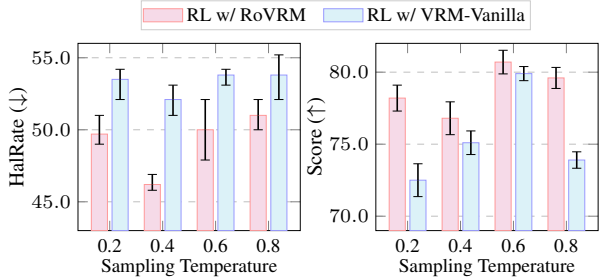
\begin{figure}
    \centering
    \scalebox{0.87}{
    \tikzstyle{every node}=[scale=1.0]
\hspace*{-0.3cm}
\begin{tikzpicture}
    \scriptsize{
      \begin{axis}[
        at={(-11.5em,8em)},
        ymajorgrids,
        grid style=dashed,
        ybar=2pt,
        enlarge x limits=0.5,
        xtick align=inside,
        height=.25\textwidth,
        width=.28\textwidth,
        bar width=1.0em,
        xlabel={Sampling Temperature},
        ylabel={Score ($\uparrow$)},
        symbolic x coords={{1}, {2}, {3}, {4}},
        xtick=data,
        nodes near coords align={vertical},
        xticklabels={{0.2},{0.4},{0.6},{0.8}},
        x tick label style={
             rotate=0,
             anchor=center,
             scale=1.2,
             xshift=-.0em,
             yshift=-0.8em
         },
        enlarge x limits=0.2,
        ylabel style={yshift=-1.3em,scale=1.3},
        xlabel style={yshift=0.8em,align=center,scale=1.2},
        ymax=82,
        ymin=69,
        ytick={70,75,...,80},
        yticklabel style={/pgf/number format/fixed,/pgf/number format/fixed zerofill,/pgf/number format/precision=1,rotate=0,scale=1.20},
        legend entries={\scalebox{1.4}{RL w/ RoVRM}, \scalebox{1.4}{RL w/ VRM-Vanilla}},
        legend columns=-1,
        legend style={
                minimum height=4ex,
                nodes={scale=0.85, transform shape},
                inner sep=.7pt,
                at={(-0.18, 1.15)},
                anchor=center,
                legend cell align=center,
			    legend plot pos=left,
                draw=black!20,
                /tikz/every even column/.append style={column sep=0.20cm}},
        ]
        \addplot+[draw=red!40,fill=myred,area legend,error bars/.cd,y dir=both,y explicit,error bar style={color=myblack,line width=0.7pt}]
        table [
            x=X, y=Y, 
            y error plus=YerrPlus, y error minus=YerrMinus
        ] {
            X   Y   YerrPlus  YerrMinus
            1       78.2    0.9     0.9
            2       76.8    1.14    1.14
            3       80.7    0.82    0.82
            4       79.6    0.73    0.73
        };
        \addplot+[draw=blue!40,fill=myblue,area legend,error bars/.cd,y dir=both,y explicit,error bar style={color=myblack,line width=0.7pt}]
        table [
            x=X, y=Y, 
            y error plus=YerrPlus, y error minus=YerrMinus
        ] {
            X   Y   YerrPlus  YerrMinus
            1       72.5    1.14    1.14
            2       75.1    0.82    0.82
            3       79.9    0.49    0.49
            4       73.9    0.57    0.57
        };
      \end{axis}

      \begin{axis}[
        at={(-30em,8em)},
        ymajorgrids,
        grid style=dashed,
        ybar=2pt,
        enlarge x limits=0.5,
        xtick align=inside,
        height=.25\textwidth,
        width=.28\textwidth,
        bar width=1.0em,
        xlabel={Sampling Temperature},
        ylabel={HalRate ($\downarrow$)},
        symbolic x coords={{1}, {2}, {3}, {4}},
        xtick=data,
        nodes near coords align={vertical},
        xticklabels={{0.2},{0.4},{0.6},{0.8}},
        x tick label style={
             rotate=0,
             anchor=center,
             scale=1.2,
             xshift=-.0em,
             yshift=-0.8em
         },
        enlarge x limits=0.2,
        ylabel style={yshift=-1.3em,scale=1.3},
        xlabel style={yshift=0.8em,align=center,scale=1.2},
        ymax=56,
        ymin=43,
        ytick={40,45,50,55,60},
        yticklabel style={/pgf/number format/fixed,/pgf/number format/fixed zerofill,/pgf/number format/precision=1,rotate=0,scale=1.20},
        ]
        \addplot+[draw=red!40,fill=myred,error bars/.cd,y dir=both,y explicit,error bar style={color=myblack,line width=0.7pt}]
        table [
            x=X, y=Y, 
            y error plus=YerrPlus, y error minus=YerrMinus
        ] {
            X   Y   YerrPlus  YerrMinus
            1       49.7    1.3     0.7
            2       46.2    0.7     0.4
            3       50.0    2.1     2.1
            4       51.0    1.1     1.0
        };
        
        \addplot+[draw=blue!40,fill=myblue,error bars/.cd,y dir=both,y explicit,error bar style={color=myblack,line width=0.7pt}]
        table [
            x=X, y=Y, 
            y error plus=YerrPlus, y error minus=YerrMinus
        ] {
            X   Y   YerrPlus  YerrMinus
            1       53.5    0.7     1.4
            2       52.1    1.0     1.1
            3       53.8    0.4     0.7
            4       53.8    1.4     1.7
        };
  
      \end{axis}  
    }
\end{tikzpicture}}
    \vspace{-2mm}
    \caption{
    Performance of the RL-trained LLaVA-1.5-7B model at temperature settings of 0.2, 0.4, 0.6, and 0.8.}
    \vspace{-2mm}
    \label{fig:diff_temperature_settings}
\end{figure}

Different temperature settings typically yield varying responses when applying an LVLM. 
To evaluate this comprehensively, we also compare the RoVRM and VRM-Vanilla under different temperature settings, as shown in Figure \ref{fig:diff_temperature_settings}. 
The results indicate that RoVRM surpasses RL's best-case performance on the MMHalBench and LLaVA-Bench benchmarks. 
This highlights the benefit of integrating textual preference data into VRM training.

\begin{table}[t]
    \centering
    \scalebox{0.74}{
    \begin{tabular}{lrccccc}
\toprule[1.1pt]
\multirow{2}{*}{Method} &
\multirow{2}{*}{Num.}  
& \multicolumn{2}{c}{AMBER}                  
& LLaVA$^\mathrm{{W}}$
& MMIns
\\ \cmidrule(l){3-4} \cmidrule(l){5-5} \cmidrule(l){6-6}
& &  Cover. $\uparrow$ & HalRate $\downarrow$ & Score $\uparrow$  & WinRate $\uparrow$  \\ \midrule
LLaVA-1.5-7B &- &50.3  &37.1 &66.7 &46.16  \\ \midrule
\multicolumn{6}{l}{\textbf{\textit{Best-of-$n$ Sampling}}} \\  \midrule
VRM-Vanilla   &- &50.8  &29.0 &73.6 &57.69 \\ \midrule
RoVRM \\
+Alpaca        &10k &50.7  &26.0 &80.6 &58.72  \\
+Helpful       &44k &50.8  &26.6 &79.3 &57.44  \\
+UltraFeedback &340k &\bf{53.2} &\bf{23.9} &82.0 &\bf{61.91}  \\
+Fusion        &394k &51.1 &26.4 &\bf{82.5} &60.71  \\ \midrule
\multicolumn{6}{l}{\textbf{\textit{Reinforcement Learning}}} \\ \midrule
VRM-Vanilla   &-   &49.1 &29.1 &72.8 &51.11 \\ \midrule
RoVRM \\
+Alpaca        &10k  &46.8 &26.5 &76.7 &56.92  \\
+Helpful       &44k  &\bf{49.5} &32.7 &75.8 &46.95  \\
+UltraFeedback &340k &48.2 &23.4 &\bf{78.3} &58.69  \\
+Fusion        &394k &44.1 &\bf{21.0} &77.9 &\bf{59.00}  \\ 
\bottomrule[1.1pt]
\end{tabular}}
    \caption{
    Performance on different textual preference datasets is evaluated using the LLaVA-1.5-7B model.
    “Num.” refers to the number of comparison preference pairs in each textual preference dataset.
    }
    \label{tab:diff_textual_datasets}
\end{table}

\begin{table}[t]
    \centering
    \scalebox{0.82}{
    \begin{tabular}{lcccc}
\toprule[1.1pt]
& \multicolumn{2}{c}{Best-of-$n$ Sampling} & \multicolumn{2}{c}{Reinforcement Learning}   \\ \cmidrule(l){2-3} \cmidrule(l){4-5}
\multirow{-2}{*}{Method} & {\begin{tabular}[c]{@{}c@{}}AMBER\\ (HalRate $\downarrow$)\end{tabular}} & {\begin{tabular}[c]{@{}c@{}}LLaVA$^\mathrm{W}$\\ (Score $\uparrow$)\end{tabular}}  & {\begin{tabular}[c]{@{}c@{}}AMBER\\ (HalRate $\downarrow$)\end{tabular}} & {\begin{tabular}[c]{@{}c@{}}LLaVA$^\mathrm{W}$\\ (Score $\uparrow$)\end{tabular}}    \\ \midrule

P1\&P2$\rightarrow$P3 &26.5    &76.9    &28.1   &70.9      \\
P1$\rightarrow$P2\&P3 &26.6    &76.7    &25.8   &72.2      \\
P1\&P2\&P3            &31.1    &74.1    &38.5   &68.0      \\
\rowcolor{gray!20}
P1$\rightarrow$P2$\rightarrow$P3 &\bf{23.9}   &\bf{82.0}    &\bf{23.4}   &\bf{78.3}      \\

\bottomrule[1.1pt]
\end{tabular}}
    \caption{Performance on the LLaVA-1.5-7B model with different preference data fusion strategies.
    \& denotes the combination of two preference datasets. $\rightarrow$ denotes a two-step training process: first, the VRM is trained with the former preference dataset at a higher learning rate, followed by the latter preference dataset at a lower learning rate. 
    P1, P2, and P3 denote the three phases depicted in Figure \ref{fig:main-image}, respectively.}
    \label{tab:data_fusion}    
\end{table}

\subsubsection{Performance on Different Textual Preference Datasets}
The performance of our RoVRM depends on the quality of the textual preference dataset. 
Therefore, in addition to UltraFeedback, we also evaluate RoVRM on other general textual preference datasets, including the Alpaca preference dataset \cite{dubois2024alpacafarm} and Helpful preference dataset \cite{bai2022training}.
Additionally, we evaluate RoVRM on a fusion of these datasets.
The preference data selection process, as described in Section \textbf{Preference Data Selection}, is applied to all datasets.
Our results show that, compared to VRM-Vanilla, utilizing each of these textual preference datasets improves the VRM's robustness, further validating the effectiveness of our three-phase progressive training and preference data selection methods.
Moreover, the results show that UltraFeedback yields the most significant performance gains, outperforming Alpaca and Helpful, likely due to its larger data scale.
Interestingly, although the Helpful dataset is also sizable, it does not outperform Alpaca. 
This discrepancy may be due to the Helpful dataset's emphasis on helpfulness, which could limit its ability to capture a broader spectrum of human preferences.
Furthermore, the results from the fusion show a slight improvement over certain benchmarks, such as LLaVA-Bench on the best-of-$n$ sampling.
However, we chose not to use the fusion in this work, as the performance gains were minimal and inconsistent across all benchmarks. 
Additionally, the fusion data incurs higher computational costs, increasing the workload of our research.

\begin{table*}[t]
    \centering
    \scalebox{0.93}{

\begin{tabular}{llllllllll}
\toprule[1.0pt]
\multicolumn{1}{l}{\multirow{2}{*}{Method}} & \multicolumn{1}{c}{\multirow{2}{*}{\#Param}} & \multicolumn{2}{c}{MMHalBench} & \multicolumn{2}{c}{Object HalBench} & \multicolumn{2}{c}{AMBER} & LLaVA$\mathrm{^W}$ & \multicolumn{1}{c}{MMIns} \\ \cmidrule(l){3-4} \cmidrule(l){5-6} \cmidrule(l){7-8} \cmidrule(l){9-9} \cmidrule(l){10-10} 

\multicolumn{1}{c}{}  & \multicolumn{1}{c}{} & Score $\uparrow$ & HalRate $\downarrow$ & Resp. $\downarrow$ & Ment. $\downarrow$ & Cover. $\uparrow$ & HalRate $\downarrow$ & Score $\uparrow$ & WinRate $\uparrow$ \\ \midrule

\multicolumn{10}{l}{\textbf{\textit{Best-of-$n$ Sampling}}}   \\ \midrule

LLaVA-NeXT-7B  & 7B         &2.59  &50.0  &16.0  &10.6  &\bf61.8  &52.8  &85.7 &90.70  \\ 
\ \ \ \ \ +VRM-Vanilla & 7B &2.61   &48.0  &14.7   &9.1  &60.9 &44.3 &99.1   &95.40    \\ \rowcolor{gray!20}
\ \ \ \ \ +RoVRM-Random & 7B &2.68 &46.3  &12.5  &8.2  &60.4  &38.6   &102.8  &97.67   \\ \rowcolor{gray!20}
\ \ \ \ \ +RoVRM & 7B  &\bf2.77  &\bf45.8  &\bf10.3  &\bf6.2  &59.2   &\bf32.8  &\bf106.1  &\bf100.00    \\ \midrule

LLaVA-NeXT-13B        & 13B    &2.71 &49.0 &14.0 &9.0 &\bf62.0 &54.4 &100.5 &92.13 \\ 
\ \ \ \ \ +VRM-Vanilla & 13B   &2.79 &47.9 &12.7 &8.3 &59.3 &44.0  &100.8 &93.33 \\ \rowcolor{gray!20}
\ \ \ \ \ +RoVRM-Random &13B &2.70 &46.5 &12.1 &7.8 &59.7 &42.5  &101.3  &95.56 \\ \rowcolor{gray!20}
\ \ \ \ \ +RoVRM & 13B    &\bf2.85 &\bf44.8 &\bf10.7 &\bf6.5 &61.2  &\bf40.2 &\bf102.4 &\bf100.00      \\\midrule

\multicolumn{10}{l}{\textbf{\textit{Reinforcement Learning}}}  \\ \midrule
LLaVA-NeXT-7B  & 7B         &2.59  &50.0  &16.0  &10.6  &\bf61.8  &52.8  &85.7 &90.70  \\ 
\ \ \ \ \ +VRM-Vanilla  & 7B    &2.68 &45.8 &13.7 &8.9 &50.9 &34.2  &87.1 &93.67 \\ \rowcolor{gray!20}
\ \ \ \ \ +RoVRM-Random &  7B   &2.70 &41.5 &11.7 &8.3 &53.8 &39.4  &89.0  &94.94  \\ \rowcolor{gray!20}
\ \ \ \ \ +RoVRM        &  7B   &\bf2.81 &\bf38.0 &\bf10.8 &\bf7.6 &55.0 &\bf30.3  &\bf91.7  &\bf96.20 \\ \midrule

LLaVA-NeXT-13B        & 13B    &2.71 &49.0 &14.0 &9.0 &\bf62.0 &54.4 &100.5 &92.13 \\ 
\ \ \ \ \ +VRM-Vanilla  & 13B &2.85 &45.2 &11.8 &8.9 &56.8 &34.7 &101.5 &94.79  \\ \rowcolor{gray!20}
\ \ \ \ \ +RoVRM-Random & 13B &2.96 &41.3 &12.1 &8.6 &55.6 &31.1 &103.0 &95.56  \\ \rowcolor{gray!20}
\ \ \ \ \ +RoVRM        & 13B &\bf3.14 &\bf37.8  &\bf10.1 &\bf5.8  &57.2 &\bf30.9&\bf106.7  &\bf96.88 \\
\bottomrule[1.0pt]
\end{tabular}}
    \caption{
    Experimental results on the LLaVA-NeXT-7B and -13B models.
    }
    \label{tab:experiments_of_llava_next}
    \vspace{-1mm}
\end{table*}

\subsubsection{Comparison with Preference Data Fusion}
Data fusion from multiple sources is a common approach to achieving knowledge transfer \cite{liu2019general}.
Intuitively, one approach to transferring preferences from text is to combine textual and visual preference datasets.
Thus, we also compare various data fusion strategies with three-phase progressive training in Table \ref{tab:data_fusion}. 
The results indicate that data fusion strategies are less effective than the three-phase progressive training, likely due to the inherent challenges of optimizing the ratio between the datasets.
This advantage in progressive training aligns with recent advancements in applying LLMs to specific tasks, where models are initially fine-tuned with general instruction data, followed by task-specific instruction data, rather than using a fusion approach \cite{panigrahi2023task,gu2021domain}.

\begin{table}[t]
    \centering
    \scalebox{0.87}{
    \begin{tabular}{lccc}
\toprule[1.1pt]
Phase             & Sample & Device & Time Cost \\ \midrule
Phase One     \\ 
\ \ \ \ Warmup Training   &5k   &1$\times$ GPU     &2.40h  \\
\ \ \ \ Gradient Feature  &58k  &8$\times$ GPU     &2.55h  \\
\ \ \ \ Optimal Transport &57k  &1$\times$ CPU     &0.55h $\sim$ 0.75h  \\  \midrule
Phase Two     \\ 
\ \ \ \ Warmup Training   &5k   &1$\times$ GPU    &2.02h  \\
\ \ \ \ Gradient Feature  &100k &8$\times$ GPU    &3.26h  \\
\ \ \ \ Optimal Transport &95k  &1$\times$ CPU    &0.55h $\sim$ 0.75h  \\
\bottomrule[1.1pt]
\end{tabular}}
    \caption{
        Time costs associated with preference data selection are reported. 
        The “Sample” column indicates the number of processed preference samples at each step.
        Due to variations in CPU processing speed, influenced by the number of active tasks or cores, we present a range of time costs derived from multiple experiments.
    }
    \vspace{-5mm}
    \label{tab:time_cost}
\end{table}

\subsubsection{Performance on Different LVLMs}
To further validate the generalizability of our RoVRM, we conduct additional experiments using the LLaVA-NeXT (as known as LLaVA-1.6) models which improve input image resolution and use an improved visual instruction tuning data \cite{liu2024llavanext}.
The experimental results, presented in Table \ref{tab:experiments_of_llava_next}, consistently demonstrate that RoVRM outperforms VRM-Vanilla in both best-of-$n$ sampling and RL training, mirroring the findings in Table \ref{tab:main_experiments}.
Notably, during RL training with RoVRM, LLaVA-NeXT-13B achieves 106.7 points on the LLaVA-Bench. 
Although LLaVA-NeXT models are already robust compared to LLaVA-1.5, RoVRM continues to improve their performance, indicating that the preferences learned by RoVRM are distinct from those captured by strong supervised fine-tuning. 
This demonstrates the effectiveness of our proposed approaches to learning preferences.

\subsubsection{Time Costs for Preference Data Selection}
Table \ref{tab:time_cost} provides the time required for our preference data selection. 
The test is performed on eight A800 GPUs and an Intel Xeon Platinum 8358P CPU. 
The results demonstrate that our approach incurs minimal time costs, ensuring its feasibility for real-world applications.

\subsubsection{Case Study}
\begin{figure*}[t]
    \centering
    \includegraphics[width=0.88\linewidth]{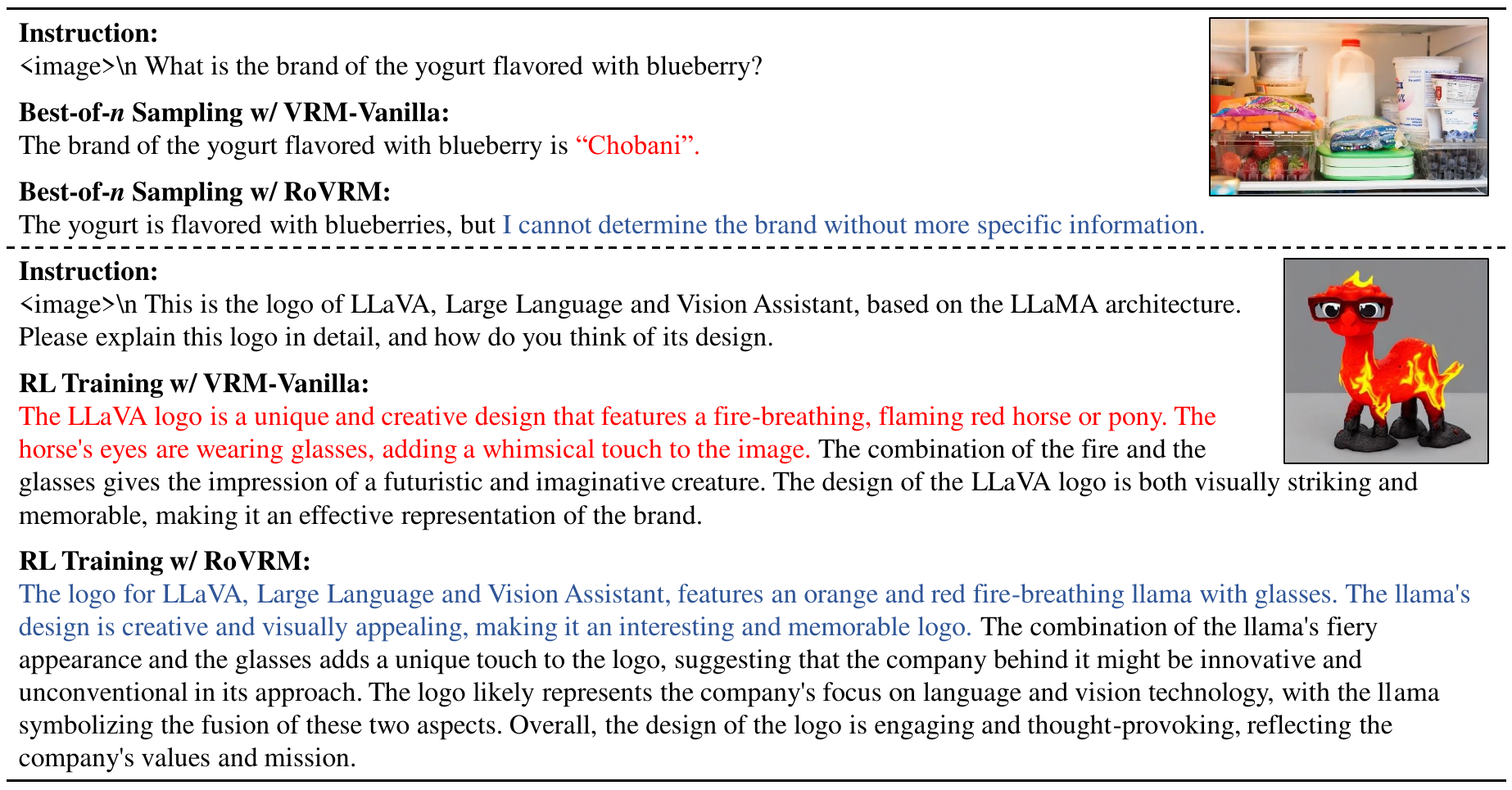}
    \caption{
    Examples where RoVRM can yield greater trustworthiness compared to VRM-Vanilla.   
    }
    \label{fig:case_study_truth}
\end{figure*}

\begin{figure*}[t]
    \centering
    \includegraphics[width=0.89\linewidth]{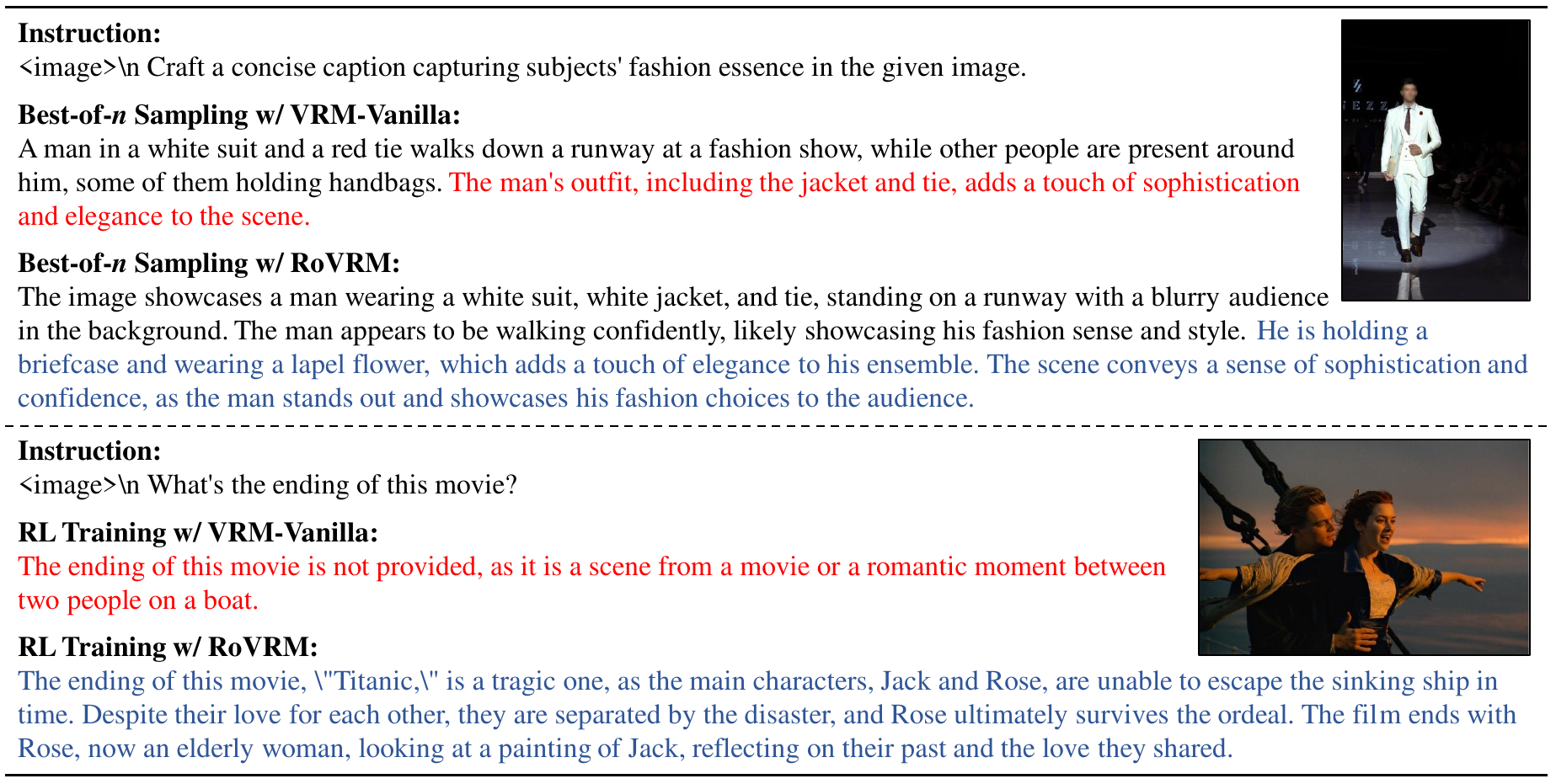}
    \caption{
    Examples where RoVRM can yield greater helpfulness compared to VRM-Vanilla.
    }
    \label{fig:case_study_helpful}
\end{figure*}

We provide several qualitative examples in Figures \ref{fig:case_study_truth} and \ref{fig:case_study_helpful} to illustrate the impact of RoVRM on aligning the LLaVA-1.5-7B model with human preferences.
These examples highlight RoVRM's ability to generate more truthful and helpful responses.

\end{document}